\documentclass[twoside]{article}
\usepackage{booktabs} 
\usepackage{multirow}
\usepackage{subfigure}
\usepackage{graphicx}
\usepackage{amssymb}
\usepackage{amsmath}
\usepackage{mystyle}
\usepackage[accepted]{aistats2019}
\usepackage{amsthm}
\usepackage{color}
\usepackage{algorithm} 
\usepackage{algorithmic}  
\usepackage[algo2e, ruled]{algorithm2e}
\usepackage{scrextend}
\usepackage{cite}

\newcommand{\GloVe}{\textsf{GloVe}}
\newcommand{\WordToVec}{\textsf{WordToVec}}

\newcommand{\myParagraph}[1]{\noindent {\bf #1} }

\renewcommand{\s}[1]{{\sffamily \small #1}}

\newcommand{\word}[1]{\texttt{#1}}
\newcommand{\wv}[2]{(\texttt{#1} - \texttt{#2})}

\usepackage[utf8]{inputenc}
\usepackage{graphicx}
\date{}

\begin{document}

\twocolumn[
\aistatstitle{Attenuating Bias in Word Vectors}

\aistatsauthor{ Sunipa Dev \And Jeff Phillips  }

\aistatsaddress{ University of Utah \And  University of Utah  } ]
\begin{abstract}
    Word vector representations are well developed tools for various NLP and Machine Learning tasks and are known to retain significant semantic and syntactic structure of languages. But they are prone to carrying and amplifying bias which can perpetrate discrimination in various applications. In this work, we explore new simple ways to detect the most stereotypically gendered words in an embedding and remove the bias from them.  We verify how names are masked carriers of gender bias and then use that as a tool to attenuate bias in embeddings. Further, we extend this property of names to show how names can be used to detect other types of bias in the embeddings such as bias based on race, ethnicity, and age. 
\end{abstract}






\section{BIAS IN WORD VECTORS}
\label{sec : intro}

Word embeddings are an increasingly popular application of neural networks wherein enormous text corpora are taken as input and words therein are mapped to a vector in some high dimensional space. Two commonly used approaches to implement this are \WordToVec~\cite{Mik1,Art5} and \GloVe~\cite{Art3}. These word vector representations estimate similarity between words based on the context of their nearby text, or to predict the likelihood of seeing words in the context of another.  Richer properties were discovered such as synonym similarity, linear word relationships, and analogies such as \word{man} : \word{woman} :: \word{king} : \word{queen}.  Their use is now standard in training complex language models.  

However, it has been observed that word embeddings are prone to express the bias inherent in the data it is extracted from \cite{debias,debias2,Caliskan183}. Further, Zhao \etal (2017) \cite{ZhaoWYOC17} and Hendricks \etal (2018) \cite{Burns2018WomenAS} show that machine learning algorithms and their output show more bias than the data they are generated from.

Word vector embeddings as used in machine learning towards applications which significantly affect people's lives, such as to assess credit~\cite{credit}, predict crime~\cite{crime}, and other emerging domains such judging loan applications and resumes for jobs or college applications.  So it is paramount that efforts are made to identify and if possible to remove bias inherent in them.  Or at least, we should attempt minimize the propagation of bias within them.  For instance, in using existing word embeddings, Bolukbasi \etal (2016) \cite{debias} demonstrated that women and men are associated with different professions, with men associated with leaderships roles and professions like doctor, programmer and women closer to professions like receptionist or nurse.  Caliskan \etal (2017) ~\cite{Caliskan183} similarly noted how word embeddings show that women are more closely associated with arts than math while it is the opposite for men. They also showed how positive and negative connotations are associated with European-American versus African-American names.

Our work simplifies, quantifies, and fine-tunes these approaches: we show that very simple linear projection of all words based on vectors captured by common names is an effective and general way to significantly reduce bias in word embeddings.  
More specifically:

\begin{itemize}
\item[1a.]
We demonstrate that simple linear projection of all word vectors along a bias direction is more effective than the Hard Debiasing of Bolukbasi \etal (2016) \cite{debias} which is more complex and also partially relies on crowd sourcing. 

\item[1b.]
We show that these results can be slightly improved by dampening the projection of words which are far from the projection distance.  

\item[2.]
We examine the bias inherent in the standard word pairs used for debiasing based on gender by randomly flipping or swapping these words in the raw text before creating the embeddings.  We show that this alone does not eliminate bias in word embeddings, corroborating that simple language modification is not as effective as repairing the word embeddings themselves.  

\item[3a.]
We show that common names with gender association (e.g., \word{john}, \word{amy}) often provides a more effective gender subspace to debias along than using gendered words (e.g., \word{he}, \word{she}).  

\item[3b.] 
We demonstrate that names carry other inherent, and sometimes unfavorable, biases associated with race, nationality, and age, which also corresponds with bias subspaces in word embeddings.  And that it is effective to use common names to establish these bias directions and remove this bias from word embeddings.  
\end{itemize}



\section{DATA AND NOTATIONS}
\label{sec:data}
We set as default the text corpus of a Wikipedia dump (\url{dumps.wikimedia.org/enwiki/latest/enwiki-latest-pages-articles.xml.bz2}) with 4.57 billion tokens and we extract a $\GloVe$ embedding from it in $D=300$ dimensions per word. We restrict the word vocabulary to the most frequent $100{,}000$ words.  We also modify the text corpus and extract embeddings from it as described later. 

So, for each word in the Vocabulary $W$, we represent the word by the vector $w_i \in \b{R}^D$ in the embedding.   The bias (e.g., gender) subspace is denoted by a set of vector $B$.  It is typically considered in this work to be a single unit vector, $v_B$ (explained in detail later).  As we will revisit, a single vector is typically sufficient, and will simplify descriptions.  However, these approaches can be generalized to a set of vectors defining a multi-dimensional subspace.

\section{HOW TO ATTENUATE BIAS}
\label{sec : methods to attenuate bias}

Given a word embedding, debiasing typically takes as input a set $\c{E} = \{E_1, E_2, \ldots, E_m\}$ of equality sets.  An equality set $E_j$ for instance can be a single pair (e.g., \{\word{man}, \word{woman}\}), but could be more words (e.g., \{\word{latina}, \word{latino}, \word{latinx}\}) that if the bias connotation (e.g, gender) is removed, then it would objectively make sense for all of them to be equal.  Our data sets will only use word pairs (as a default the ones in Table \ref{tbl:word-pairs}), and we will describe them as such hereafter for simpler descriptions.  In particular, we will represent each $E_j$ as a set of two vectors $e_i^+, e_i^- \in \b{R}^D$.  

Given such a set $\c{E}$ of equality sets, the bias vector $v_B$ can be formed as follows~\cite{debias}.  For each $E_j = \{e_j^+, e_j^-\}$ create a vector $\vec{e}_i = e_i^+ - e_i^-$ between the pairs.  Stack these to form a matrix $Q = [\vec{e}_1\; \vec{e}_2\; \ldots\; \vec{e}_m]$, and let $v_B$ be the top singular vector of $Q$.  We revisit how to create such a bias direction in Section \ref{sec:gender-subspace}.  

\begin{table}[]
	\small
    \centering
    \begin{tabular}{c}
\hline
    \{\word{man},\word{woman}\}, \{\word{son},\word{daughter}\}, \{\word{he},\word{she}\}, \{\word{his},\word{her}\},\\
    \{\word{male},\word{female}\},
   \{\word{boy},\word{girl}\}, \{\word{himself},\word{herself}\}, \\ \{\word{guy},\word{gal}\},
   \{\word{father},\word{mother}\}, \{\word{john},\word{mary}\} \\ \hline
    \end{tabular}
    \caption{Gendered Word Pairs}
    \label{tbl:word-pairs}
\end{table}
\normalsize

Now given a word vector $w \in W$, we can project it to its component along this bias direction $v_B$ as 
\[
\pi_B(w) = \langle w, v_B \rangle v_B.
\]

\subsection{Existing Method : Hard Debiasing}
The most notable advance towards debiasing embeddings along the gender direction has been by Bolukbasi \etal (2016) \cite{debias} in their algorithm called Hard Debiasing ($HD$). 
%
It takes a set of words desired to be neutralized, $\{w_1, w_2, \ldots, w_n\} = W_N \subset W$, a unit bias subspace vector $v_B$, and a set of equality sets $E_1, E_2, \ldots, E_m$. 

First, words $\{w_1, w_2, \ldots, w_n\} \in W_N$ are projected orthogonal to the bias direction and normalized  
\[
w_i' = \frac{w_i - w_B}{||w_i - w_B||}.  
\]


Second, it corrects the locations of the vectors in the equality sets.  Let $\mu_j = \frac{1}{|E|} \sum_{e \in E_j} e$ be the mean of an equality set, and $\mu = \frac{1}{m} \sum_{j=1}^m \mu_j$ be the mean of of equality set means.  Let $\nu_j = \mu - \mu_j$ be the offset of a particular equality set from the mean.  
Now each $e \in E_j$ in each equality set $E_j$ is first centered using their average and then neutralized as
\[
e' = \nu_j + \sqrt{1-\|\nu_j\|^2}\frac{\pi_B(e) - v_B}{\|\pi_B(e)- v_B\|}.  
\] 
Intuitively $\nu_j$ quantifies the amount words in each equality set $E_j$ differ from each other in directions apart from the gender direction. This is used to center the words in each of these sets.

This renders word pairs such as \word{man} and \word{woman} as equidistant from the neutral words $w_i'$ with each word of the pair being centralized and moved to a position opposite the other in the space. This can filter out properties either word gained by being used in some other context, like \word{mankind} or \word{humans} for the word \word{man}.

The word set $W_N = \{w_1,w _2, \ldots, w_n\} \subset W$ which is debiased is obtained in two steps.  
First it seeds some words as definitionally gendered via crowd sourcing and using dictionary definitions; the complement -- ones not selected in this step -- are set as neutral. 
Next, using this seeding an SVM is trained and used to predict among all $W$ the set of other biased $W_B$ or neutral words $W_N$. 
This set $W_N$ is taken as desired to be neutral and is debiased.  
Thus not all words $W$ in the vocabulary are debiased in this procedure, only a select set chosen via crowd-sourcing and definitions, and its extrapolation.  
Also the word vectors in the equality sets are also handled separately.  
This makes this approach not a fully automatic way to debias the vector embedding.

\subsection{Alternate and Simple Methods}
\label{sec : projection}

We next present some simple alternatives to HD which are simple and fully automatic.  
These all assume a bias direction $v_B$.  

\myParagraph{Subtraction.  }
As a simple baseline, for \emph{all} word vectors $w$  subtract the gender direction $v_B$ from $w$:
\[
w' = w - v_B. 
\]

\myParagraph{Linear Projection.  }
A better baseline is to project \emph{all} words $w \in W$ orthogonally to the bias vector $v_B$.  \[
w' = w - \pi_B(w) = w - \langle w, v_B \rangle v_B.  
\]
This enforces that the updated set $W' = \{w' \mid w \in W\}$ has no component along $v_B$, and hence the resulting span is only $D-1$ dimensions.  Reducing the total dimension from say $300$ to $299$ should have minimal effects of expressiveness or generalizability of the word vector embeddings. 

Bolukbasi \etal \cite{debias} apply this same step to a dictionary definition based extrapolation and crowd-source-chosen set of word pairs $W_N \subset W$.  We quantify in Section \ref{sec : tests} that this single universal projection step debiases better than HD.

For example, consider the bias as gender, and the equality set with words \word{man} and \word{woman}.  Linear projection will subtract from their word embeddings the proportion that were along the gender direction $v_B$ learned from a larger set of equality pairs.   It will make them close-by but not exactly equal.   
The word \word{man} is used in many extra senses than the word \word{woman}; it is used to refer to humankind, to a person in general, and in expressions like ``oh man''.  
In contrast a simpler word pair with fewer word senses, like \wv{he}{she} and \wv{him}{her}, we can expect them to be almost at identical positions in the vector space after debiasing, implying their synonymity. 

Thus, this approach uniformly reduces the component of the word along the bias direction without compromising on the differences that words (and word pairs) have.  

%

\subsection{Partial Projection}
\label{sec : variants}
A potential issue with the simple approaches is that they can significantly change some embedded words which are definitionally biased (e.g., the neutral words $W_B$ described by Bolukbasi \etal~\cite{debias}).  \emph{[[We note that this may not *actually* be a problem (see Section \ref{sec : tests}); the change may only be associated with the bias, so removing it would then not change the meaning of those words in any way except the ones we want to avoid.]]} However, these intuitively should be words which have correlation with the bias vector, but also are far in the orthogonal direction.  In this section we explore how to automatically attenuate the effect of the projection on these words.   

This stems from the observation that given a bias direction, the words which are most extreme in this direction (have the largest dot product) sometimes have a reasonable biased context, but some do not.  These ``false positives'' may be large normed vectors which also happen to have a component in the bias direction.  


\begin{figure}
\begin{center}
\includegraphics[width=0.9\linewidth]{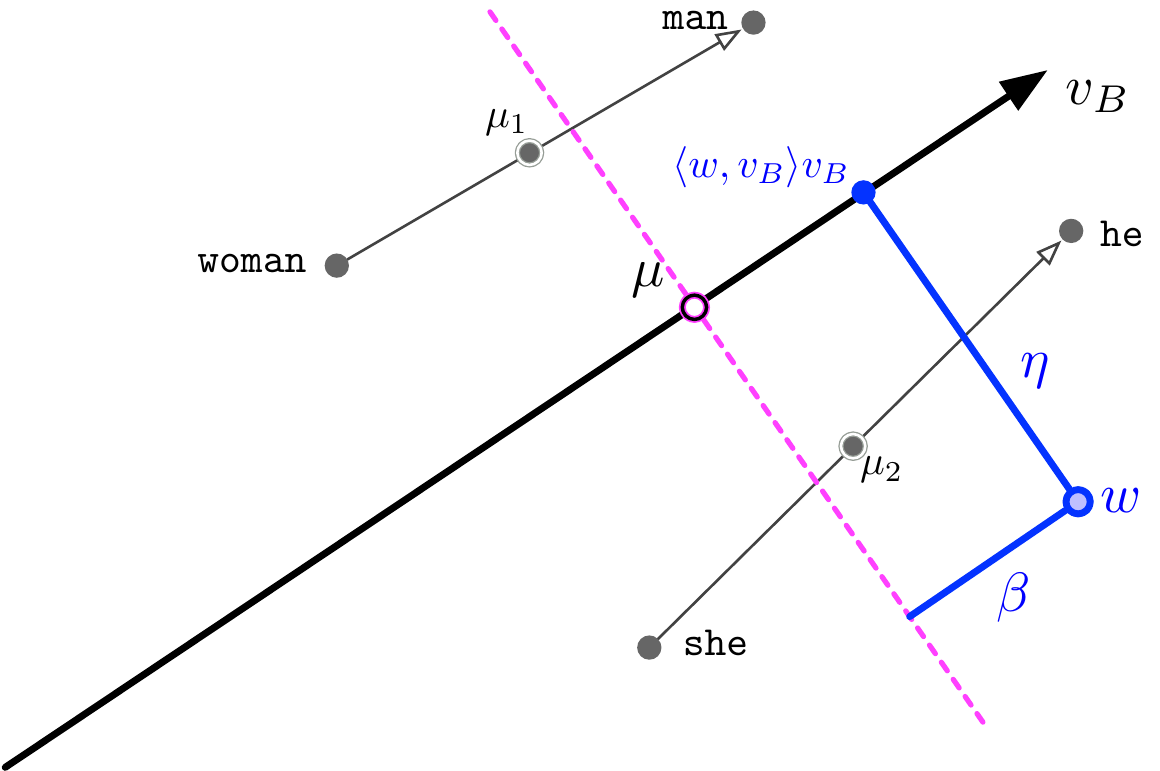}
\end{center}

\vspace{-3mm}
\caption{\label{fig:eta-beta}Illustration of $\eta$ and $\beta$ for word vector $w$.}
\end{figure}

We start with a bias direction $v_B$ and mean $\mu$ derived from equality pairs (defined the same way as in context of HD).  
Now given a word vector $w$ we decompose it into key values along two components, illustrated in Figure \ref{fig:eta-beta}.  
First, we write its bias component as 
\[
\beta(w) = \langle w, v_B \rangle - \langle \mu, v_B \rangle.
\]
This is the difference of $w$ from $\mu$ when both are projected onto the bias direction $v_B$.  

Second, we write a (residual) orthogonal component 
\[
r(w) = w - \langle w, v_B \rangle v_B.  
\]
Let $\eta(w) = \|r(w)\|$ be its value. 
It is the orthogonal distance from the bias vector $v_B$; recall we chose $v_B$ to pass through the origin, so the choice of $\mu$ does not affect this distance.  

Now we will maintain the orthogonal component ($r(w)$, which is in a subspace spanned by $D-1$ out of $D$ dimensions) but adjust the bias component $\beta(w)$ to make it closer to $\mu$.  But the adjustment will depend on the magnitude $\eta(w)$.  
As a default we set 
\[
w' = \mu + r(w)
\]
so all word vectors retain their orthogonal component, but have a fixed and constant bias term.  
This is functionally equivalent to the Linear Projection approach; the only difference is that instead of having a $0$ magnitude along $v_B$ (and the orthogonal part unchanged), it instead has a magnitude of constant $\mu$ along $v_B$ (and the orthogonal part still unchanged).  This adds a constant to every inner product, and a constant offset to any linear projection or classifier.  If we are required to work with normalize vectors (we do not recommend this as the vector length captures veracity information~
about its embedding), we can simple set $w' = r(w)/\|r(w)\|$.

Given this set-up, we now propose three modifications.  In each set 
\[
w' = \mu + r(w) + \beta \cdot f_i(\eta(w)) \cdot v_B
\]
were $f_i$ for $i = \{1,2,3\}$ is a function of only the orthogonal value $\eta(w)$.  For the default case $f(\eta) = 0$
\begin{align*}
f_1(\eta) &= \sigma^2/(\eta+1)^2
\\
f_2(\eta) &= \exp(-\eta^2/\sigma^2)
\\
f_3(\eta) &= \max(0,\sigma/2\eta) 
\end{align*}
Here $\sigma$ is a hyperparameter that controls the importance of $\eta$; in Section \ref{app:sigma} we show that we can just set $\sigma=1$.  x

\begin{figure}
	\includegraphics[width = 0.42 \textwidth]{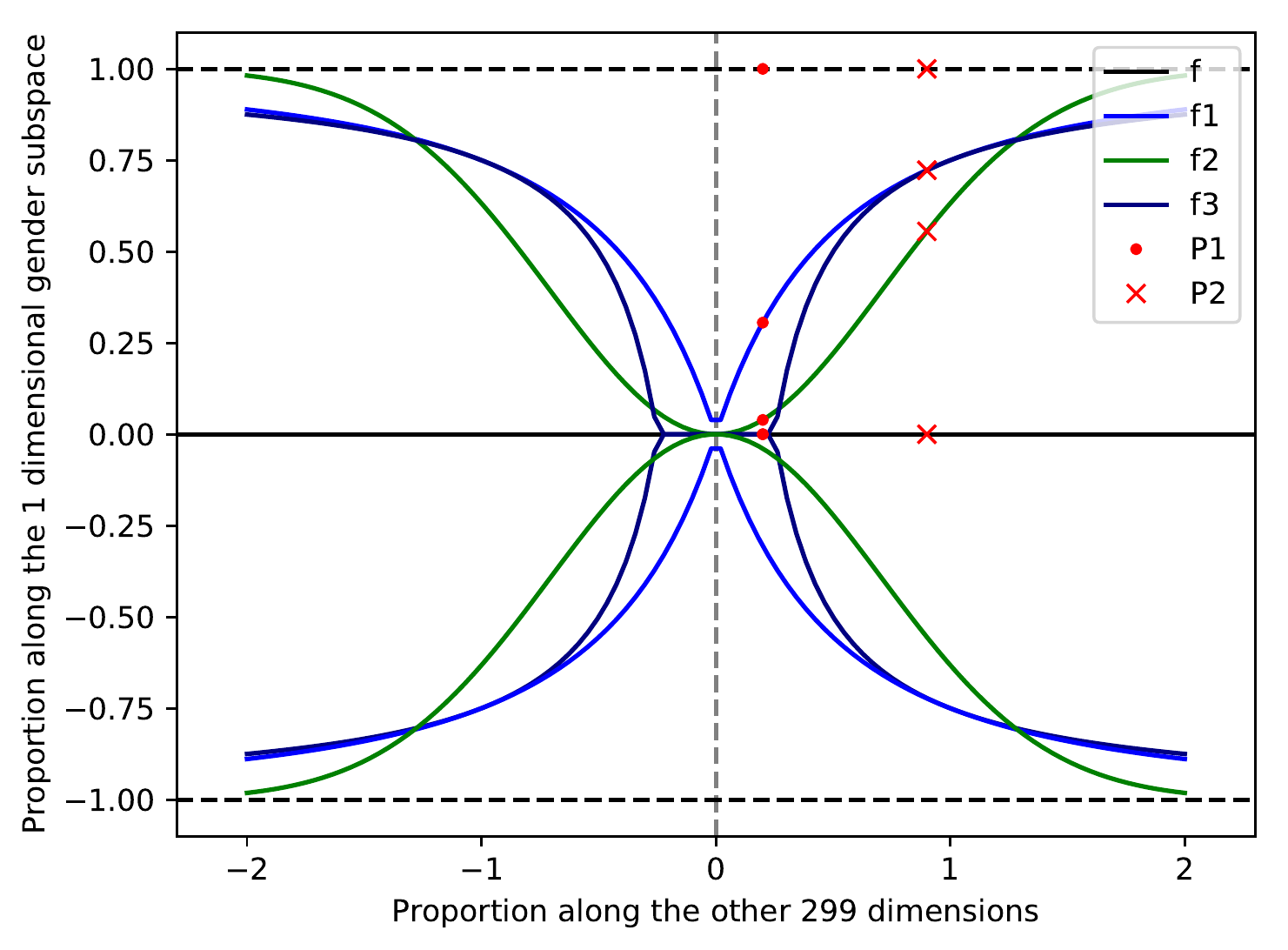}
	\small
	\caption{The gendered region as per the three variations of projection. Both points P1 and P2 have a dot product of 1.0 initially with the gender subspace. But their orthogonal distance to it differs, as expressed by their dot product with the other 299 dimensions. \label{fig : variations}}
	\normalsize
\end{figure}

In Figure \ref{fig : variations} we see the regions of the $(\eta,\beta)$-space that the functions $f$, $f_1$ and $f_2$ consider gendered. $f$ projects all points onto the $y = \mu$ line. But variants $f_1$, $f_2$, and $f_3$ are represented by curves that dampen the bias reduction to different degrees as $\eta$ increases.  Points P1 and P2 have the same dot products with the bias direction but different dot products along the other $D-1$ dimensions. We can observe the effects of each dampening function as $\eta$ increases from P1 to P2.

\subsection{SETTING $\sigma=1$.}
\label{app:sigma}
To complete the damping functions $f_1$, $f_2$, and $f_3$, we need a value $\sigma$.  If $\sigma$ is larger, then more word vectors have bias completely removed; if $\sigma$ is smaller, than more words are unaffected by the projection.  
The goal is that words $S$ which are likely to carry a bias connotation to have little damping (small $f_i$ values) and words $T$ which are unlikely to carry a bias connotation to have more damping (large $f_i$ values -- they are not moved much).  

Given sets $S$ and $T$, we can define a gain function
\[
\gamma_{i,\rho}(\sigma) =  \sum_{s \in S} \beta(s) (1-f_i(\eta(s))) -  \rho \sum_{t \in T} \beta(t) (1-f_i(\eta(t))),
\]
with a regularization term $\rho$. 
The gain $\gamma$ is large when most bias words in $S$ have very little damping (small $f_i$, large $1-f_i$), and the opposite is true for the neutral words in $T$.  We want the neutral words to have large $f_i$ and hence small $1-f_i$, so they do not change much.

To define the gain function, we need sets $S$ and $T$; we do so with the bias of interest as gender.  The biased set $S$ is chosen among a set of $1000$ popular names in $W$ which (based on babynamewizard.com and and SSN databases~\cite{nameWizard,ssn}) are strongly associated with a gender.  The neutral set $T$ is chosen as the most frequent $1000$ words from $W$, after filtering out obviously gendered words like names \word{man} and \word{he}.  We also omit occupation words like \word{doctor} and others which may carry unintentional gender bias (these are words we would like to automatically de-bias).  The neutral set may not be perfectly non-gendered, but it provides reasonable approximation of all non-gendered words.

We find for an array of choices for $\rho$ (we tried $\rho =1$, $\rho=10$, and $\rho=100$), the value $\sigma=1$ approximately maximizes the gain function $\gamma_{i,\rho}(\sigma)$ for each $i \in \{1,2,3\}$.  So for hereafter we fix $\sigma=1$.  

Although these sets $S$ and $T$ play a role somewhat similar to the crowd-sourced sets $W_B$ and $W_N$ from HD that we hoped to avoid, the role here is much reduced.  This is just to verify that a choice of $\sigma=1$ is reasonable, and otherwise they are not used.   

\begin{figure}[h]
	\includegraphics[width=0.24\linewidth]{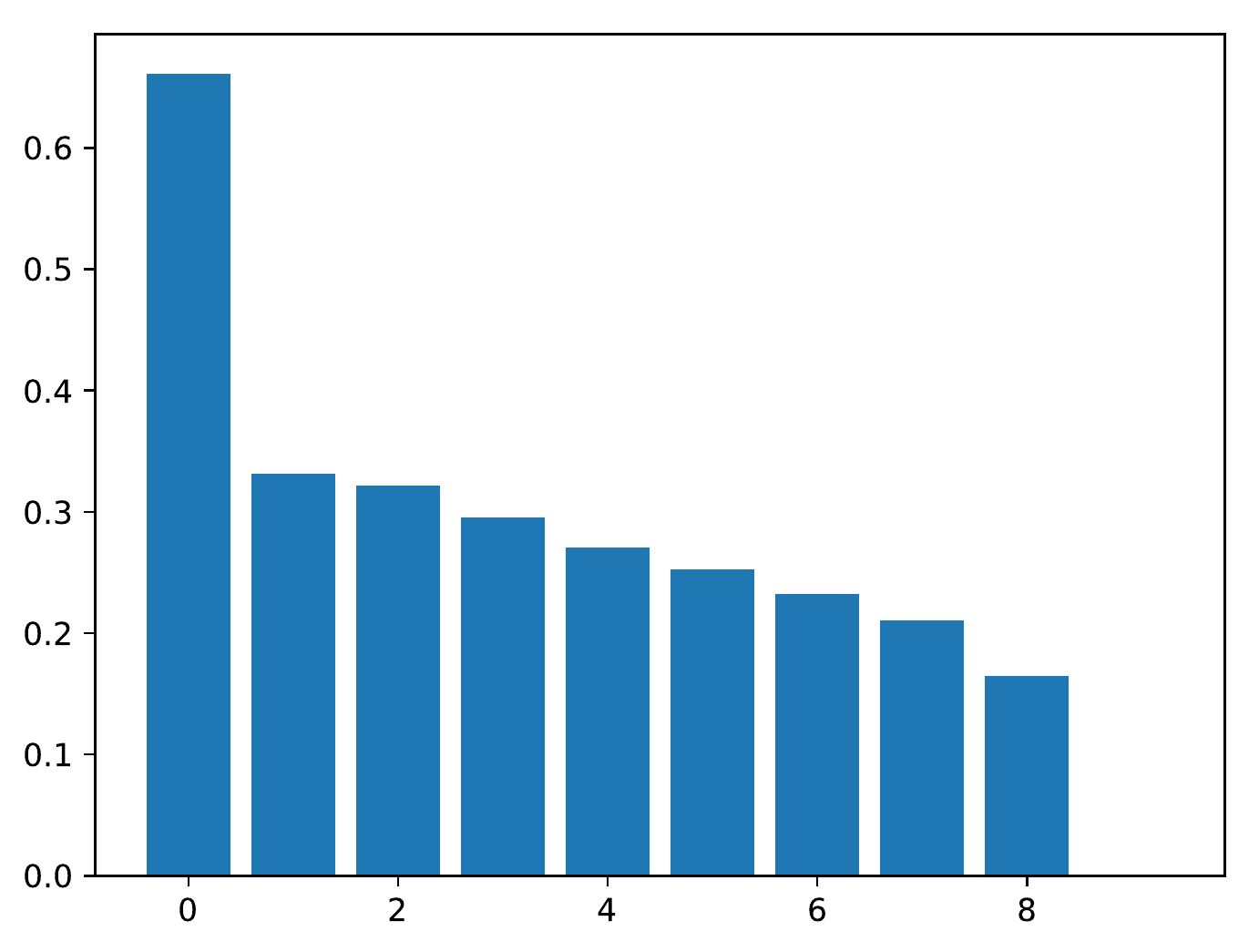}
	\includegraphics[width=0.24\linewidth]{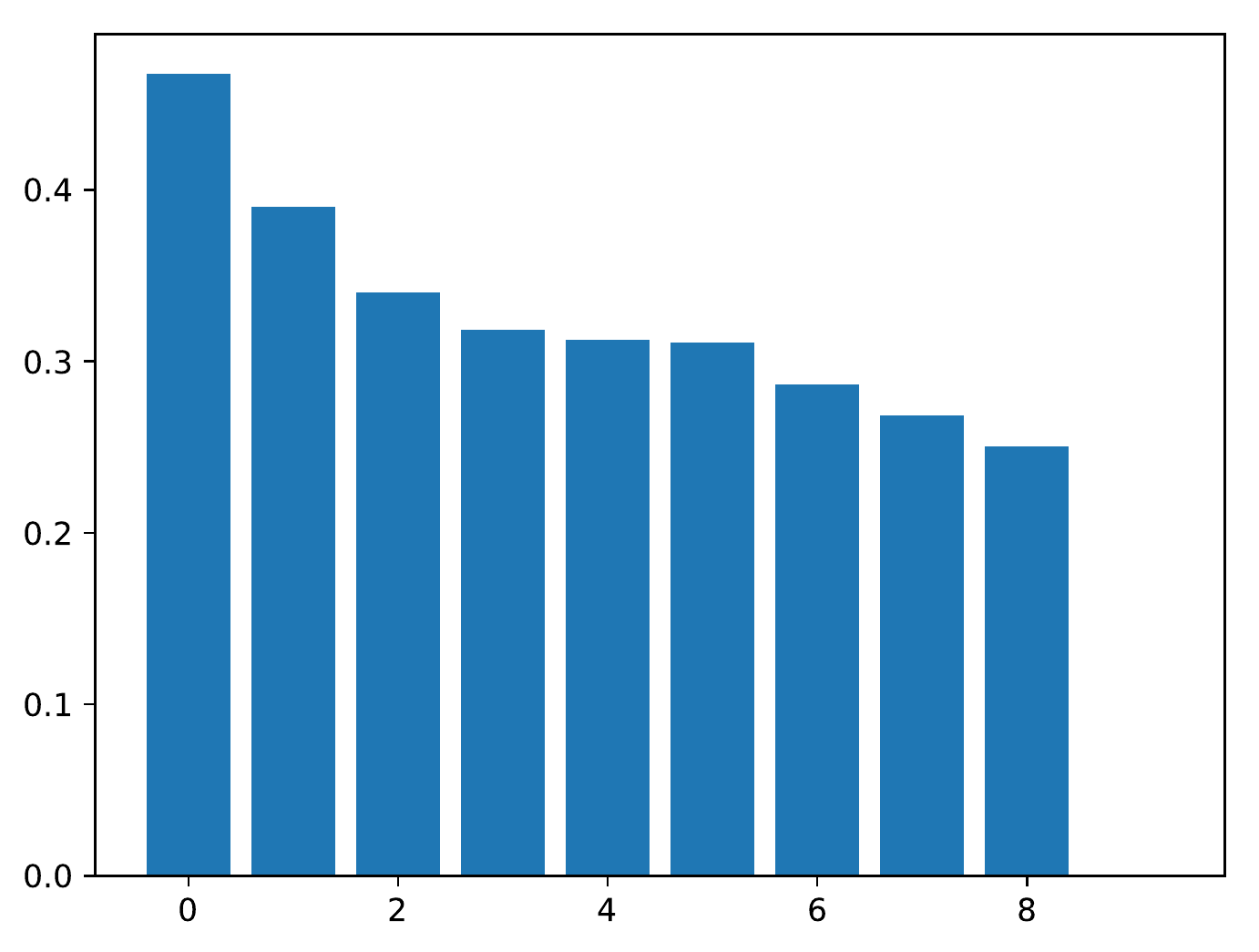}
	\includegraphics[width=0.24\linewidth]{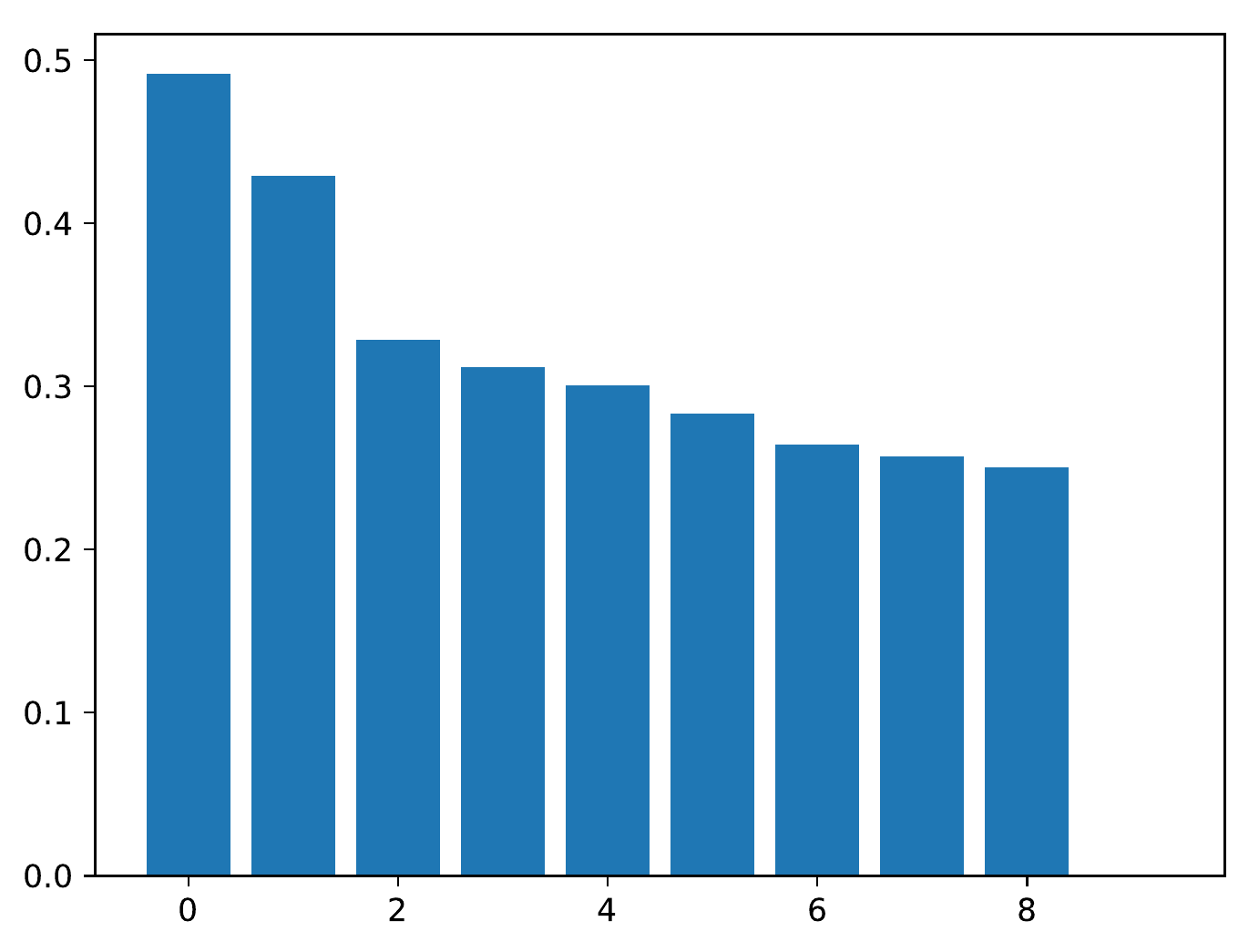}
	\includegraphics[width=0.24\linewidth]{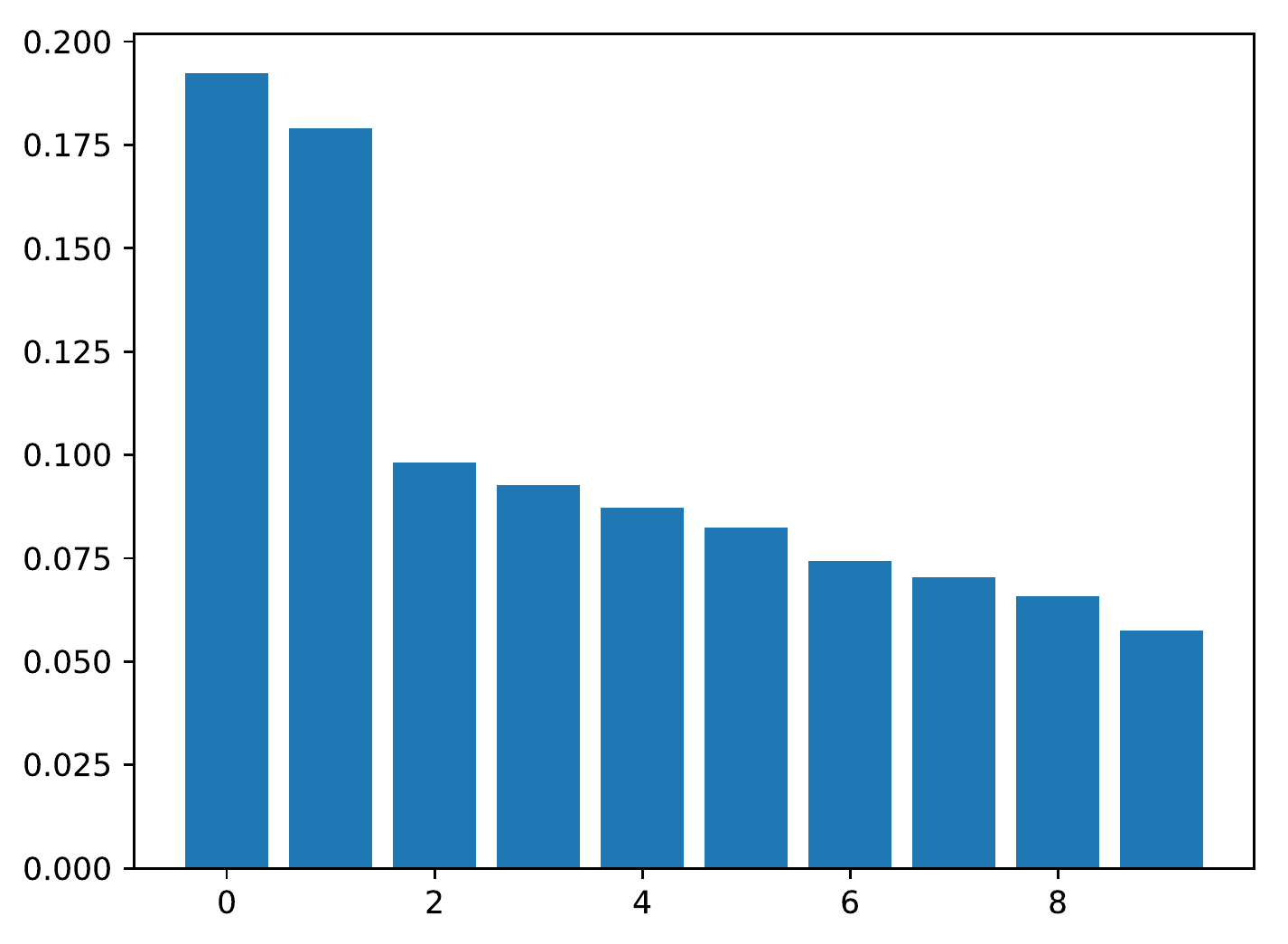}
	\\
	\phantom{123} $0.00$ \hspace{13mm} $0.50$ \hspace{12mm} $0.75$ \hspace{12mm} $1.00$
	
	\vspace{-3mm}
	\small 
	\caption{\label{fig : flipping} Fractional singular values for avg male - female words (as per Table \ref{tbl:word-pairs}) after flipping with probability (from left to right) $0.0$ (the original data set), $0.5$, $0.75$,  and $1.0$.}
\end{figure}
\normalsize

\subsection{Flipping the Raw Text}
Since the embeddings preserve inner products of the data from which it is drawn, we explore if we can make the data itself gender unbiased and then observe how that change shows up in the embedding.  Unbiasing a textual corpus completely can be very intricate and complicated since there are a many (sometimes implicit) gender indicators in text. 
Nonetheless, we propose a simple way of neutralizing bias in textual data by using word pairs $E_1, E_2, \ldots E_m$; in particular, when we observe in raw text on part of a word part, we randomly flip it to the other pair.  For instance for gendered word pairs (e.g., \wv{he}{she}) in a string ``\word{he} was a doctor'' we may flip to ``\word{she} was a doctor.''

We implement this procedure over the entire input raw text, and try various probabilities of flipping each observed word, focusing on probabilities $0.5$, $0.75$ and $1.00$.  The first $0.5$-flip probability makes each element of a word pair equally likely.  The last $1.00$-flip probability reverses the roles of those word pairs, and $0.75$-flip probability does something in between.  
We perform this set of experiments on the default Wikipedia data set and switch between word pairs (say \word{man} $\rightarrow$ \word{woman}, \word{she} $\rightarrow$ \word{he}, etc), from a list larger that Table \ref{tbl : Gendered Words} consisting of 75 word pairs; see Supplementary Material \ref{app : flipping}.  

We observe how the proportion along the principal component changes with this flipping in Figure \ref{fig : flipping}.  
We see that flipping with $0.5$ somewhat dampens the difference between the different principal components.  
On the other hand flipping with probability $1.0$ (and to a lesser extent $0.75$) exacerbates the gender components rather than dampening it. Now there are two components significantly larger than the others.  This indicates this flipping is only addressing part of the explicit bias, but missing some implicit bias, and these effects are now muddled.  

We list some gender biased analogies in the default embedding and how they change with each of the methods described in this section in Table \ref{tbl : analogies flipping}.

\begin{table*}[]
	\centering
	\small
	\caption{What analogies look like before and after damping gender by different methods discussed : hard Debiaisng, flipping words in text corpus, subtraction and projection}
	\label{tbl : analogies flipping}
	\hspace{-1.5cm}
	\begin{tabular}{c|c|c|ccc|c|c}
		analogy head & original & HD & \multicolumn{3}{c|}{flipping} & subtraction  & projection \\
		\cline{4-6}
		& & & 0.5 & 0.75 & 1.0 & & \\\hline
		\word{man} : \word{woman} :: \word{doctor} : &  \word{nurse} & \word{surgeon}  & \word{dr} & \word{dr} &  \word{medicine} & \word{physician} & \word{physician}\\
		\word{man} : \word{woman} :: \word{footballer} :  & \word{politician} & \word{striker} & \word{midfielder} & \word{goalkeeper} & \word{striker} & \word{politician} & \word{midfielder}\\
		
		\word{he} : \word{she} :: \word{strong} : & \word{weak} & \word{stronger} & \word{weak} & \word{strongly} &  \word{many} & \word{well} & \word{stronger}\\
		\word{he} : \word{she} :: \word{captain} : & \word{mrs} & \word{lieutenant} &  \word{lieutenant} &  \word{colonel} & \word{colonel} & \word{lieutenant} & \word{lieutenant} \\
		\word{john} : \word{mary} :: \word{doctor} : & \word{nurse} & \word{physician} & \word{medicine} & \word{surgeon} & \word{nurse} & \word{father} & \word{physician}\\ 
	\end{tabular}
\normalsize	
\end{table*}

\section{THE BIAS SUBSPACE}
\label{sec:gender-subspace}

%

We explore ways of detecting and defining the bias subspace $v_B$ and recovering the most gendered words in the embedding.  Recall as default, we use $v_B$ as the top singular vector of the matrix defined by stacking vectors $\vec{e}_i = e_i^+ - e_i^-$ of biased word pairs.  
We primarily focus on gendered bias, using words in Table \ref{tbl:word-pairs}, and show later how to effectively extend to other biases.  We discuss this in detail in Supplementary Material \ref{app : gender dir}.

\paragraph{Most gendered words.}
The dot product, $\langle v_B, w \rangle$ of the word vectors $w$ with the gender subspace  $v_B$ is a good indicator of how gendered a word is. 
The magnitude of the dot product tells us of the length along the gender subspace and the sign tells us whether it is more female or male.   
Some of the words denoted as most gendered are listed in Table \ref{tbl : Gendered Words}. 



\begin{table}
	\small
	\centering
    \caption{Some of the most gendered words in default embedding; and most gendered adjectives and occupation words.}  
\label{tbl : Gendered Words}
    \begin{tabular}{c c c c }
         \multicolumn{4}{c}{Gendered Words}\\ \hline 
         \word{miss} & \word{herself} & \word{forefather} & \word{himself}\\
         \word{maid} & \word{heroine} & \word{nephew} & \word{congressman} \\
         \word{motherhood} & \word{jessica} & \word{zahir} & \word{suceeded}\\
         \word{adriana} & \word{seductive} & \word{him} & \word{sir}\\
         
    \end{tabular}

    \begin{tabular}{c c}
	Female Adjectives & Male Adjectives  \\ \hline 
	\word{glamorous} & \word{strong}\\
	\word{diva} & \word{muscular}\\
	\word{shimmery} & \word{powerful} \\
	\word{beautiful} & \word{fast}\\
\end{tabular}

\begin{tabular}{c c}
	Female Occupations & Male Occupations  \\ \hline 
	\word{nurse} & \word{soldier}\\
	\word{maid} & \word{captain}\\
	\word{housewife} & \word{officer}\\
	\word{prostitute} & \word{footballer}\\
\end{tabular}

\end{table}
\normalsize





 
\subsection{Bias Direction using Names}
\label{detecting bias with names}

When listing gendered words by $|\langle v_B, w \rangle|$, we observe that many gendered words are names. This indicates the potential to use names as an alternative (and potentially in a more general way) to bootstrap finding the gender direction. 

\begin{figure}
	\centering
	\includegraphics[width=0.15\textwidth, height = 1.5cm]{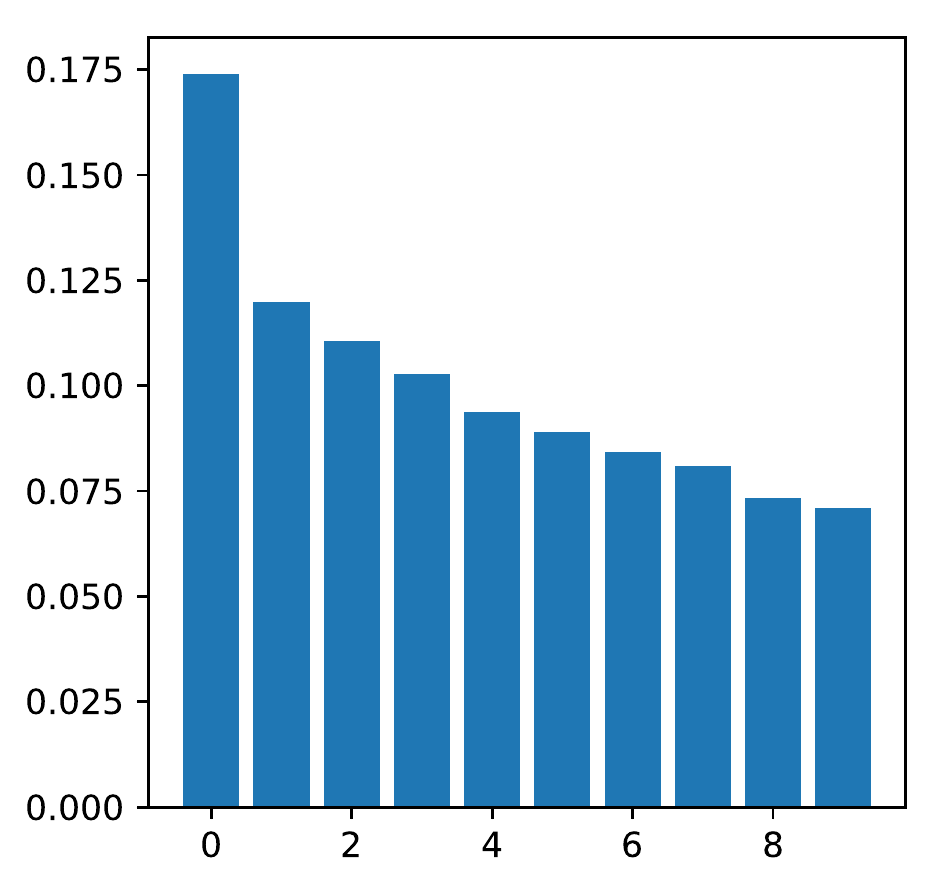}
	\includegraphics[width=0.15\textwidth, height= 1.5cm]{gender1.pdf}
	\small
	\caption{Proportion of singular values along principal directions (left) using names as indicators, and (right) using word pairs from Table \ref{tbl:word-pairs} as indicators}
	\label{fig : names}
\end{figure}

From the top 100K words, we extract the 10 most common male $\{m_1, m_2, \ldots, m_{10}\}$ and female $\{s_1,s_2,\ldots, s_{10}\}$ names which are not used in ambiguous ways (e.g., not the name \word{hope} which could also refer to the sentiment). 
We pair these 10 names from each category (male, female) randomly and compute the SVD as before.  
We observe in \ref{fig : names} that the fractional singular values show a similar pattern as with the list of correctly gendered word pairs like \wv{man}{woman}, \wv{he}{she}, etc.  
\normalsize
But this way of pairing names is quite imprecise. These names are not `opposites' of each other in the sense that word pairs are. So, we modify how we compute $v_B$ now so that we can better use names to detect the bias in the embedding. The following method gives us this advantage where we do not necessarily need word pairs or equality sets as in Bolukbasi \etal \cite{debias}.

Our gender direction is calculated as,
\[
v_{B,\textsf{names}} = \frac{s - m}{\|s - m\|},
\]
where $s = \frac{1}{10}\sum_{i} s_i$ and 
$m = \frac{1}{10}\sum_{i} m_i$.  


Using the default Wikipedia dataset, we found that this is a good approximator of the gender subspace defined by the first right singular vector calculated using gendered words from Table \ref{tbl:word-pairs}; there dot product is $0.809$.
We find similar large dot product scores for other datasets too. 

Here too we collect all the most gendered words as per the gender direction $v_{B,\textsf{names}}$ determined by these names. Most gendered words returned are similar as using the default $v_B$, like occupational words, adjectives, and synonyms for each gender. We find names to express similar classification of words along male - female vectors with \word{homemaker} more female and \word{policeman} being more male. We illustrate this in more detail in Table \ref{tbl : occ names}. 

\begin{table}
	\small
	\caption{Gendered occupations as observed in word embeddings using names as the gender direction indicator }
	\begin{tabular}{c c | c c}
		Female Occ & Male Occ & Female* Occ & Male* Occ \\ \hline 
		\word{nurse} & \word{captain} & \word{policeman} & \word{policeman}\\
		\word{maid} & \word{cop} & \word{detective} & \word{cop}\\
		\word{actress} & \word{boss} & \word{character} & \word{character}\\
		\word{housewife} & \word{officer} & \word{cop} & \word{assassin}\\
		\word{dancer} & \word{actor} & \word{assassin} & \word{bodyguard}\\
		\word{nun} & \word{scientist} & \word{actor} & \word{waiter}\\
		\word{waitress} & \word{gangster} & \word{waiter} & \word{actor}\\
		\word{scientist} & \word{trucker} & \word{butler} & \word{detective}\\
	\end{tabular}
	\normalsize
	\label{tbl : occ names}
\end{table}
Using that direction, we debias by linear projection. There is a similar shift in analogy results. We see a few examples in Table \ref{tbl : analogies}.
\begin{table*}[]
	\small
	\caption{What analogies look like before and after removing the gender direction using names}
	\centering
	\begin{tabular}{c|ccc}
		analogy head & original & subtraction & projection \\ \hline
		\word{man} : \word{woman} :: \word{doctor} :  &  \word{nurse} & \word{physician} & \word{physician} \\
		\word{man} : \word{woman} :: \word{footballer} : & \word{politician} & \word{politician}  & \word{midfielder}\\
		
		\word{he} : \word{she} :: \word{strong} : &  \word{weak} & \word{very} & \word{stronger} \\
		\word{he} : \word{she} :: \word{captain} : &  \word{mrs} & \word{lieutenant} & \word{lieutenant}\\
		\word{john} : \word{mary} :: \word{doctor} : &  \word{nurse} & \word{dr} &  \word{dr} \\
	\end{tabular}
	\normalsize
	\label{tbl : analogies}
\end{table*}

\section{QUANTIFYING BIAS}
\label{sec : tests}
In this section we develop new measures to quantify how much bias has been removed from an embedding, and evaluate the various techniques we have developed for doing so.

As one measure, we use the Word Embedding Association Test (WEAT) test developed by Caliskan \etal (2017) \cite{Caliskan183} as analogous to the IAT tests to evaluate the association of male and female gendered words with two categories of target words: career oriented words versus family oriented words.  We detail WEAT and list the exact words used (as in \cite{Caliskan183}) in Supplementary Material \ref{app : weat}; smaller values are better.

Bolukbasi \etal \cite{debias} evaluated embedding bias use a crowsourced judgement of whether an analogy produced by an embedding is biased or not.  Our goal was to avoid crowd sourcing, so we propose two more automatic tests to qualitatively and uniformly evaluate an embedding for the presence of gender bias.


\paragraph{Embedding Coherence Test (ECT). }

A way to evaluate how the neutralization technique affects the embedding is to evaluate how the nearest neighbors change for (a) gendered pairs of words $\c{E}$ and (b) indirect-bias-affected words such as those associated with sports or occupational words (e.g., \word{football}, \word{captain}, \word{doctor}).  
We use the gendered word pairs in Table \ref{tbl:word-pairs} for $\c{E}$ and the professions list $P = \{p_1, p_2, \ldots, p_k\}$ as proposed and used by Bolukbasi \etal \url{https://github.com/tolga-b/debiaswe} (see also Supplementary Material \ref{sec : occupation words}) to represent (b).

\begin{itemize}
\item[S1:] 
For all word pair $\{e_j^+, e_j^-\} = E_j \in \c{E}$ we compute two means $m = \frac{1}{|\c{E}|} \sum_{E_j \in \c{E}} e_j^+$ and $s = \frac{1}{|\c{E}|} \sum_{E_j \in \c{E}} e_j^-$.   We find the cosine similarity of both $m$ and $s$ to all words $p_i \in P$.  This creates two vectors $u_m, u_s \in \b{R}^k$.  

\item[S2:] 
We transform these similarity vectors to replace each coordinate by its rank order, and compute the Spearman Coefficient (in $[-1,1]$, larger is better) between the rank order of the similarities to words in $P$. 
\end{itemize}

Thus, here, we care about the order in which the words in $P$ occur as neighbors to each word pair rather than the exact distance. The exact distance between each word pair would depend on the usage of each word and thus on all the different dimensions other than the gender subspace too. But the order being relatively the same, as determined using Spearman Coefficient would indicate the dampening of bias in the gender direction (i.e., if \word{doctor} by profession is the 2nd closest of all professions to both \word{man} and \word{woman}, then the embedding has a dampened bias for the word \word{doctor} in the gender direction). 
Neutralization should ideally bring the Spearman coefficient towards 1.

\paragraph{Embedding Quality Test (EQT).  }
 
The demonstration by Bolukbasi \etal \cite{debias} about the skewed gender roles in embeddings using analogies is what we try to quantify in this test. We attempt to quantify the improvement in analogies with respect to bias in the embeddings. 

We use the same sets $\c{E}$ and $P$ as in the ECT test.  
However, for each profession $p_i \in P$ we create a list $S_i$ of their plurals and synonyms from WordNet on NLTK~\cite{nltk}.

\begin{itemize}
\item[S1:] 
For each word pair $\{e_j^+, e_j^-\} = E_j \in \c{E}$, and each occupation word $p_i \in P$, we test if the analogy  
$e_j^+ : e_j^- :: p_i$ 
returns a word from $S_i$. If yes, we set $Q(E_j,p_i) = 1$, and $Q(E_j,p_i) = 0$ otherwise.  

\item[S2:]
Return the average value across all combinations
$\frac{1}{|\c{E}|}\frac{1}{k} \sum_{E_j \in \c{E}} \sum_{p_i \in P} Q(E_j, p_i)$.  
\end{itemize}

The scores for EQT are typically much smaller than for ECT.  We explain two reasons for this.  

First, EQT does not check for if the analogy makes relative sense, biased or otherwise. So, ``\word{man} : \word{woman} :: \word{doctor} : \word{nurse}'' 
is as wrong as 
``\word{man} : \word{woman} :: \word{doctor} : \word{chair}." 
This pushes the score down.

Second, synonyms in each set $s_i$ as returned by WordNet \cite{WN} on the Natural Language Toolkit, NLTK \cite{nltk} do not always contain all possible variants of the word. For example, the words \word{psychiatrist} and \word{psychologist} can be seen as analogous for our purposes here but linguistically are removed enough that WordNet does not put them as synonyms together.  Hence, even after debiasing, if the analogy returns 
``\word{man} : \word{woman} :: \word{psychiatrist} : \word{psychologist}`` 
S1 returns 0. Further, since the data also has several misspelt words, \word{archeologist} is not recognized as a synonym or alternative for the word \word{archaeologist}. 
For this too S1 returns a 0.

The first caveat can be side-stepped by restricting the pool of words we search over for the analogous word to be from list $P$. But it is debatable if an embedding should be penalized equally for returning both nurse or chair for the analogy 
``\word{man} : \word{woman} :: \word{doctor} : ?'' 

This measures the quality of analogies, with better quality having a score closer to $1$.

 \begin{table*}[]
 	\small
\caption{Performance on ECT, EQT and WEAT by the different debiasing methods; and performance on standard similarity and analogy tests.}

    \centering

    \begin{tabular}{c|c|c|ccc|cc|cc}
    	\hline
        analogy head & original & HD & \multicolumn{3}{c|}{flipping} & \multicolumn{2}{c|}{subtraction}  & \multicolumn{2}{c}{projection} \\
       \cline{4-10}
         & & & 0.5 & 0.75 & 1.0 & word pairs & names & word pairs & names\\
           \hline 
           ECT (word pairs) & 0.798 & 0.917  & 0.983  & 0.984 & 0.683   & 0.963  & 0.936  & 0.996 & 0.943 \\
           ECT (names) & 0.832 & 0.968  & 0.714  & 0.662 & 0.587  & 0.923  & 0.966  & 0.935 & 0.999 \\
           EQT  & 0.128 & 0.145  & 0.131  &0.098 & 0.085  & 0.268 & 0.236  & 0.283 & 0.291\\
           WEAT & 1.623 & 1.221 & 1.164 & 1.09 & 1.03  & 1.427  & 1.440 & 1.233 & 1.219\\
         \hline
     		WSim & 0.637 & 0.537 & 0.567 & 0.537 & 0.536 & 0.627 & 0.636 & 0.627 & 0.629\\
		Simlex & 0.324 & 0.314 & 0.317 & 0.314 & 0.264 & 0.302 & 0.312 & 0.321 & 0.321\\
		Google Analogy & 0.623 & 0.561 & 0.565 & 0.561 & 0.321 & 0.538 & 0.565 & 0.565 & 0.584 \\ \hline 
    \end{tabular}

    \label{tbl : Test 2 - analogies}
\end{table*}
\normalsize

\paragraph{Evaluating embeddings.}
We mainly run $4$ methods to evaluate our methods WEAT, EQT, and two variants of ECT:
 ECT (word pairs) uses $\c{E}$ defined by words in Table \ref{tbl:word-pairs} and 
 ECT (names) which uses vectors $m$ and $s$ derived by gendered names. 
 
We observe in Table \ref{tbl : Test 2 - analogies} that the ECT score increases for all methods in comparison to the non-debiased (the original) word embedding; the exception is flipping with $1.0$ probability score for ECT (word pairs) and all flipping variants for ECT (names).  Flipping does nothing to affect the names, so it is not surprising that it does not improve this score; further indicating that it is challenging to directly fix bias in raw text before creating embeddings.  
Moreover, HD has the lowest score (of $0.917$) whereas projection obtains scores of $0.996$ (with $v_B$) and $0.943$ (with $v_{B,\textsf{names}}$).  

EQT is a more challenging test, and the original embedding only achieves a score of $0.128$, and HD only obtains $0.145$ (that is $12-15\%$ of occupation words have their related word as nearest neighbor).  On the other hand, projection increases this percentage to $28.3\%$ (using $v_B$) and $29.1\%$ (using $v_{B,\textsf{names}}$).  Even subtraction does nearly as well at between $23-27\%$.  Generally, the subtraction always performs slightly worse than projection.  

For the WEAT test, the original data has a score of $1.623$, and this is decreased the most by all forms of flipping, down to about $1.1$.  HD and projection do about the same with HD obtaining a score of $1.221$ and projection obtaining $1.219$ (with $v_{B,\textsf{names}}$) and $1.234$ (with $v_B$); values closer to 0 are better (See Supplementary Material \ref{app : weat}).    

In the bottom of Table \ref{tbl : Test 2 - analogies} we also run these approaches on standard similarity and analogy tests for evaluating the quality of embeddings.  We use cosine similarity~\cite{Lev2} on 
WordSimilarity-353 (WSim, 353 word pairs) \cite{wsim} and 
SimLex-999 (Simlex, 999 word pairs) \cite{simlex}, 
each of which evaluates a Spearman coefficient (larger is better).  We also use   
the Google Analogy Dataset using the function 3COSADD \cite{Lev1} which takes in three words which for a part of the analogy and returns the 4th word which fits the analogy the best. 

%

We observe (as expected) that all that debiasing approaches reduce these scores.  The largest decrease in scores (between $1\%$ and $10\%$) is almost always from HD.  Flipping at $0.5$ rate is comparable to HD.  And simple linear projection decreases the least (usually only about $1\%$, except on analogies where it is $7\%$ (with $v_B$) or $5\%$ (with $v_{B,\textsf{names}}$).

In Table \ref{tbl:damping} we also evaluate the damping mechanisms defined by $f_1$, $f_2$, and $f_3$, using $v_B$.  These are very comparable to simple linear projection (represented by $f$).  The scores for ECT, EQT, and WEAT are all about the same as simple linear projection, usually slightly worse.  

\begin{table}
	\small
	\centering
	\caption{\label{tbl:damping} Performance of damped linear projection using word pairs.
}
	\begin{tabular}{c|ccccc}
		\hline

		Tests & f & $f_1$ & $f_2$ & $f_3$\\ \hline
		ECT & 0.996 &0.994 & 0.995 & 0.997\\
		EQT & 0.283& 0.280& 0.292 & 0.287\\
		WEAT & 1.233 & 1.253 & 1.245 & 1.241\\ \hline
		WSim & 0.627 & 0.628 & 0.627 & 0.627\\
		Simlex & 0.321 & 0.324 & 0.324 & 0.324\\
		Google Analogy & 0.565 & 0.571 & 0.569 & 0.569\\				
		\hline
	\end{tabular}
\end{table}
\normalsize

While ECT, EQT and WEAT scores are in a similar range for all of $f$, $f_1$, $f_2$, and $f_3$; the dampened approaches $f_1$, $f_2$, and $f_3$ performs better on the Google Analogy test.  This test set is devoid of bias and is made up of syntactic and semantic analogies. So, a score closer to that of the original, biased embedding, tells us that more structure has been retained by $f_1$, $f_2$ and $f_3$.  Overall, any of these approaches could be used if a user wants to debias while retaining as much structure as possible, but otherwise linear projection (or $f$) is roughly as good as these dampened approaches.  

\section{DETECTING OTHER BIAS USING NAMES}
We saw so far how projection combined with finding the gender direction using names works well and works as well as projection combined with finding the gender direction using word pairs. 

We explore here a way of extending this approach to detect other kinds of bias where we cannot necessarily find good word pairs to indicate a direction, like Table \ref{tbl:word-pairs} for gender, but where names are known to belong to certain protected demographic groups. For example, there is a divide between names that different racial groups tend to use more. Caliskan \etal \cite{Caliskan183} use a list of names that are more African-American (AA) versus names that are more European-American (EA) for their analysis of bias. There are similar lists of names that are distinctly and commonly used by different ethnic, racial (e.g.,  Asian, African-American) and even religious (for e.g., Islamic) groups.  


We first try this with two common demographic group divides : Hispanic / European-American and African-American / European-American. 

\hangindent = 0.3cm
\myParagraph{Hispanic and European-American names.} 
Even though we begin with most commonly used Hispanic (H) names (Supplementary Material \ref{app : word lists}), this is tricky as not all names occur as much as European American names and are thus not as well embedded. We use the frequencies from the dataset to guide us in selecting commonly used names that are also most frequent in the Wikipedia dataset.  Using the same method as Section \ref{detecting bias with names}, we determine the direction, $v_{B,\textsf{names}}$, which encodes this racial difference and find the words most commonly aligned with it.  Other Hispanic and European-American names are the closest words. But other words like, \word{latino} or \word{hispanic} also appear to be close, which affirms that we are capturing the right subspace.

\begin{figure}
\includegraphics[width=.98\linewidth]{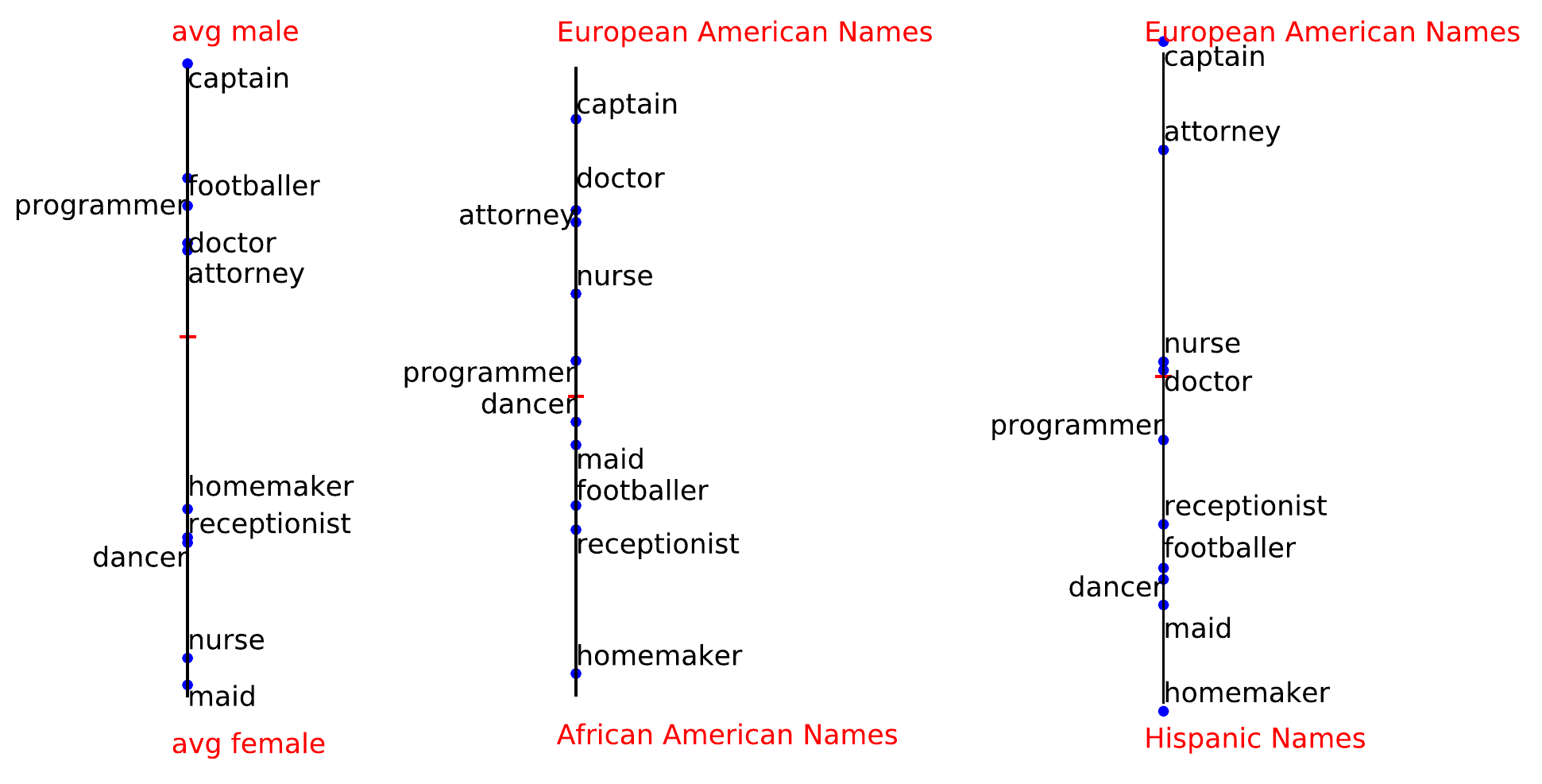}
\vspace{-4mm}
\caption{\label{fig : hispanic} Gender and racial bias in the embedding}
\end{figure}

\hangindent = 0.3cm
\myParagraph{African-American and European-American names.} 
We see a similar trend when we use African-American names and European-American names (Figure \ref{fig : hispanic}). We use the African-American names used by Caliskan \etal (2017) \cite{Caliskan183}. We determine the bias direction by using method in Section \ref{detecting bias with names}. 

We plot in Figure \ref{fig : hispanic} a few occupation words along the axes defined by H-EA and AA-EA bias directions, and compare them with those along the male-female axis.  The embedding is different among the groups, and likely still generally more subordinate-biased towards Hispanic and African-American names as it was for female.  Although \word{footballer} is more Hispanic than European-American, while \word{maid} is more neutral in the racial bias setting than the gender setting.  
We see this pattern repeated across embeddings and datasets (see Supplementary Material \ref{app : bias in different embeddings}).


When we switch the type of bias, we also end up finding different patterns in the embeddings. In the case of both of these racial directions, there is a the split in not just occupation words but other words that are detected as highly associated with the bias subspace. It shows up foremost among the closest words of the subspace of the bias. Here, we find words like \word{drugs} and \word{illegal} close to the H-EA direction while, close to the AA-EA direction, we retrieve several slang words used to refer to African-Americans.  
These word associations with each racial group can be detected by the WEAT tests (lower means less bias) using positive and negative words as demonstrated by Caliskan \etal (2017) \cite{Caliskan183}.  
We evaluate using the WEAT test before and after linear projection debiasing in Table \ref{tbl : weat other bias}.  For each of these tests, we use half of the names in each category for finding the bias direction and the other half for WEAT testing. This selection is done arbitrarily and the scores are averaged over 3 such selections.

\begin{table}[]
    \centering
    \small
     \caption{ WEAT positive-negative test scores before and after debiasing}
    \label{tbl : weat other bias}
    \begin{tabular}{c|cc}
    	\hline
        & Before Debiasing & After Debiasing  \\
        \hline
        EA-AA & 1.803 & 0.425 \\
        EA-H & 1.461& 0.480\\ 
		Youth-Aged & 0.915 & 0.704\\ \hline
    \end{tabular}

\vspace{-4mm}

\end{table}
\normalsize

More qualitatively, as a result of the dampening of bias, we see that biased words like other names belonging to these specific demographic groups, slang words, colloquial terms like \word{latinos} are removed from the closest 10\% words. This is beneficial since the distinguishability of demographic characteristics based on names is what shows up in these different ways like in occupational or financial bias. 

\begin{figure}
	\centering
	
	\includegraphics[width=0.4\textwidth, height = 0.18 \textheight]{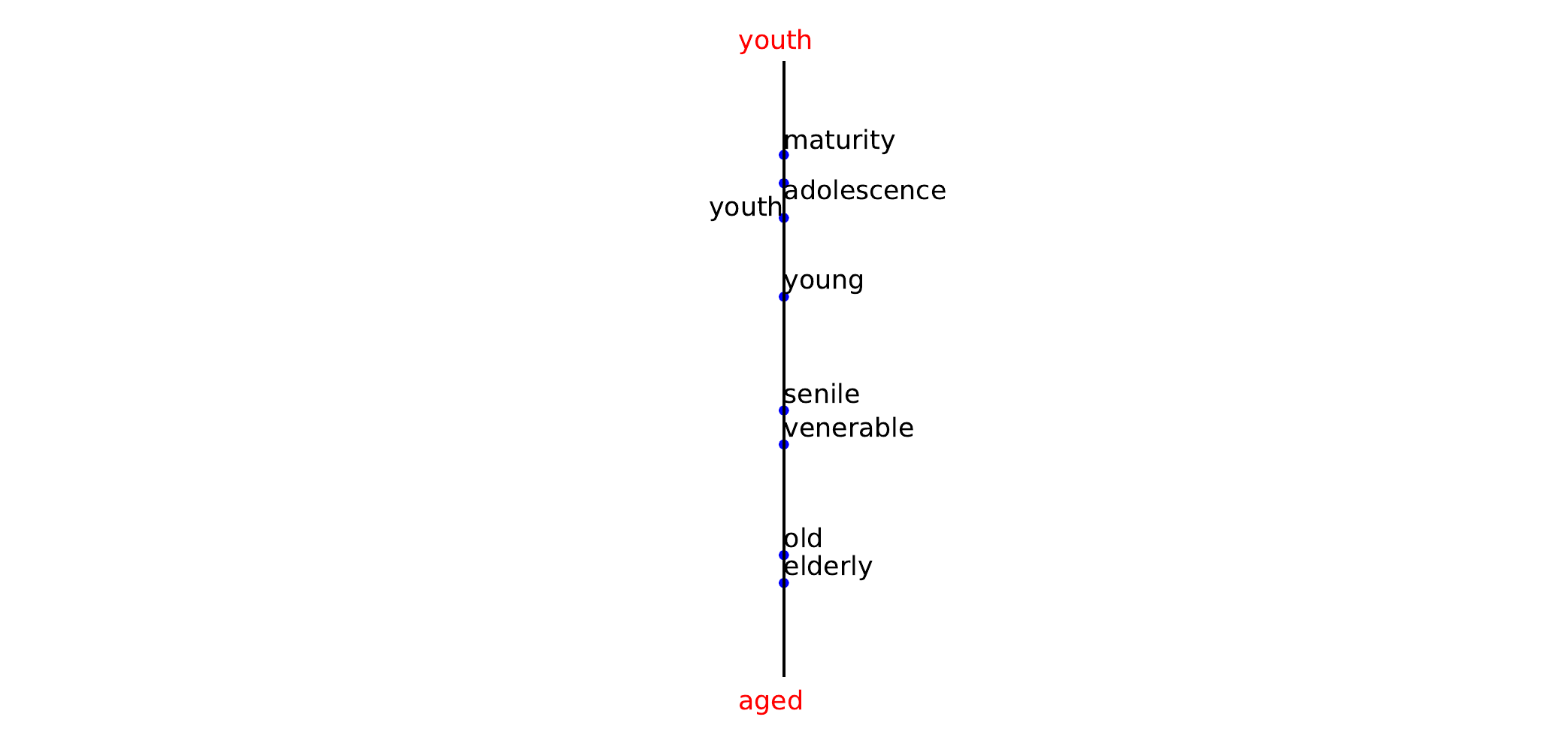}

	\caption{Detecting Age with Names : a plot of age related terms along names from different centuries.}
	\label{fig : age}
	
\end{figure}

\paragraph{Age-associated names.} 
We observed that names can be masked carriers of age too. Using the database for names through time~\cite{ssn} and extracting the most common names from early 1900s as compared to late 1900s and early 2000s, we find a correlation between these names (see Supplementary Material) and age related words. In Figure \ref{fig : age}, we see
a clear correlation between age and names. Bias in this case does not show up in professions as clearly as in gender but in terms of association with positive and negative words \cite{Caliskan183}. We again evaluate using a WEAT test in Table \ref{tbl : weat other bias}, the bias before and after debiasing the embedding. 

\section{DISCUSSION}

Different types of bias exist in textual data. Some are easier to detect and evaluate. Some are harder to find suitable and frequent indicators for and thus, to dampen. Gendered word pairs and gendered names are frequent enough in textual data to allow us to successfully measure it in different ways and project the word embeddings away from the subspace occupied by gender. Other types of bias don't always have a list of word pairs to fall back on to do the same. But using names, as we see here, we can measure and detect the different biases anyway and then project the embedding away from. In this work we also see how a weighted variant of propection removes bias while retaining best the inherent structure of the word embedding. 



\bibliographystyle{plain}
\bibliography{icpbib}

\newpage

\pagenumbering{arabic}
\begin{center}
{\sffamily \LARGE
Supplementary material for: \\
Attenuating Bias in Word Embeddings}
\\
\end{center}

\appendix

\section{Bias in different embeddings}
\label{app : bias in different embeddings}

\begin{figure*}
\centering
   (a)
    \includegraphics[width=0.2\textwidth,height=0.2\textwidth]{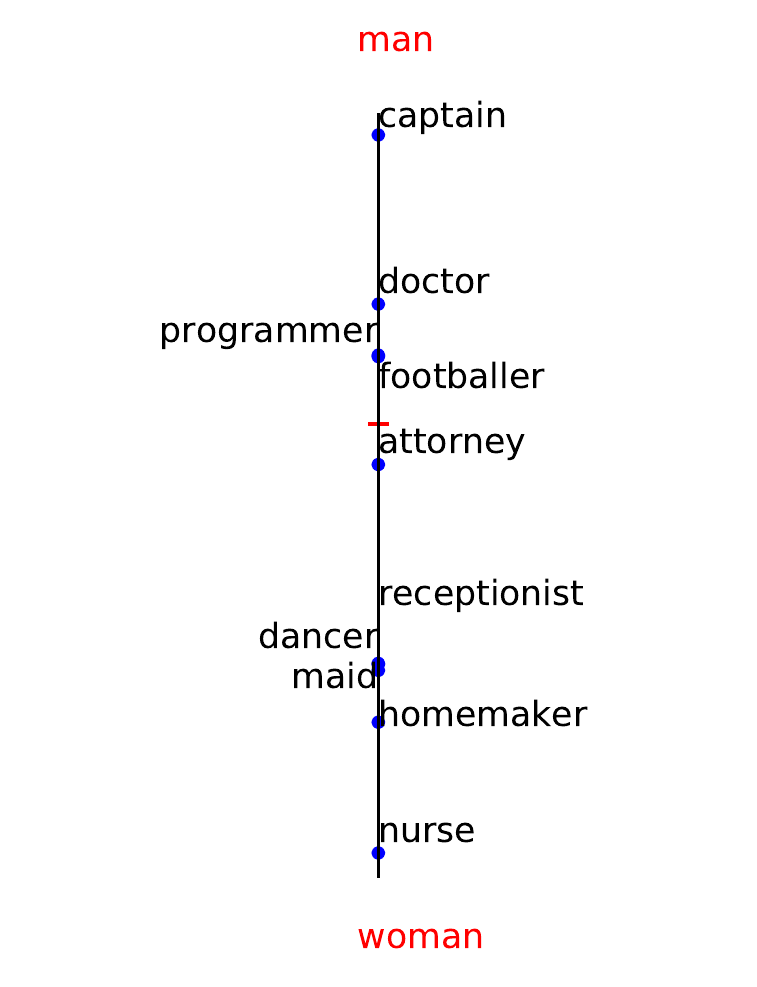}
   \includegraphics[width=0.2\textwidth, height=0.2\textwidth]{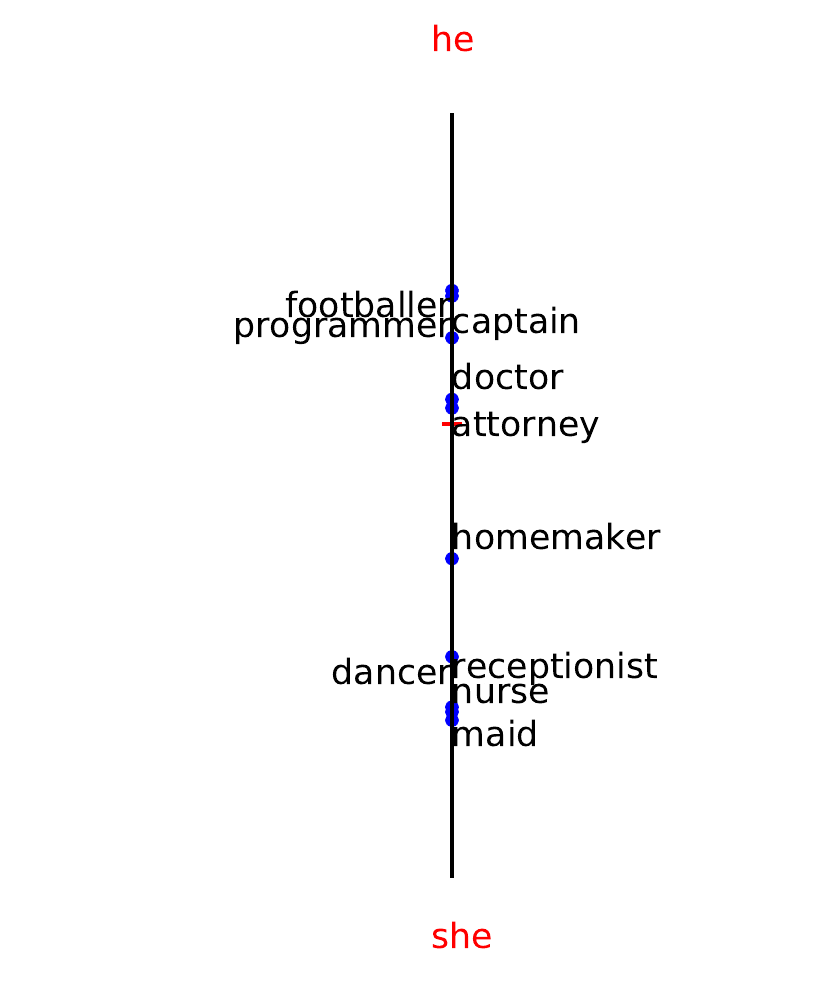}
    \includegraphics[width=0.2\textwidth, height=0.2\textwidth]{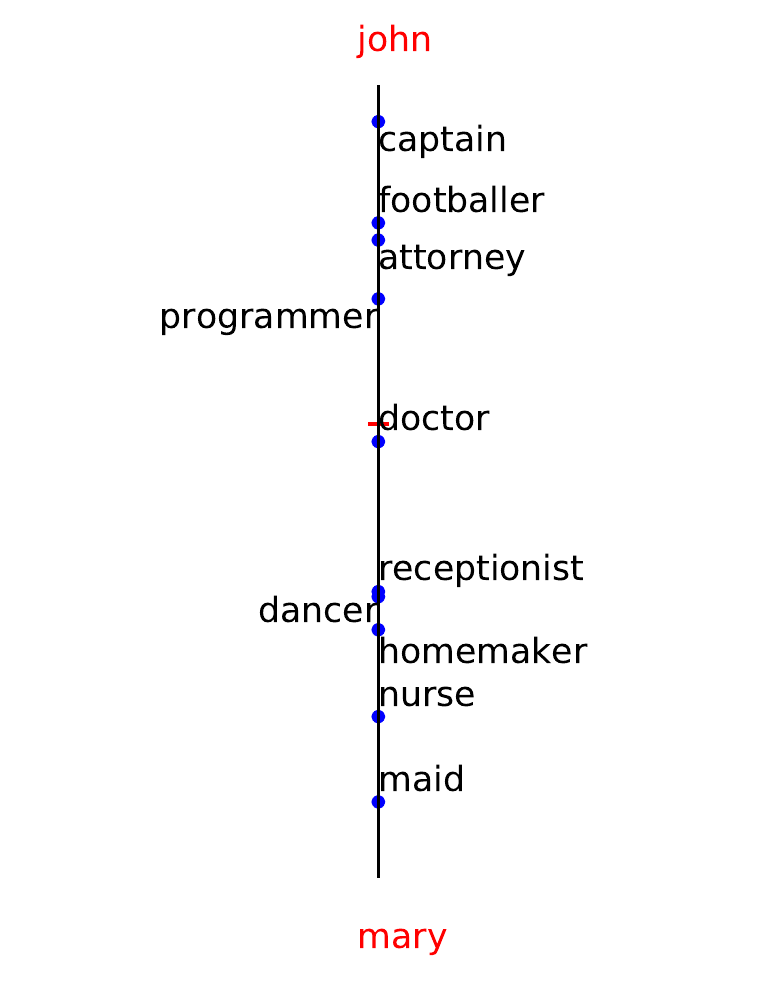}
    \includegraphics[width=0.2\textwidth, height=0.2\textwidth]{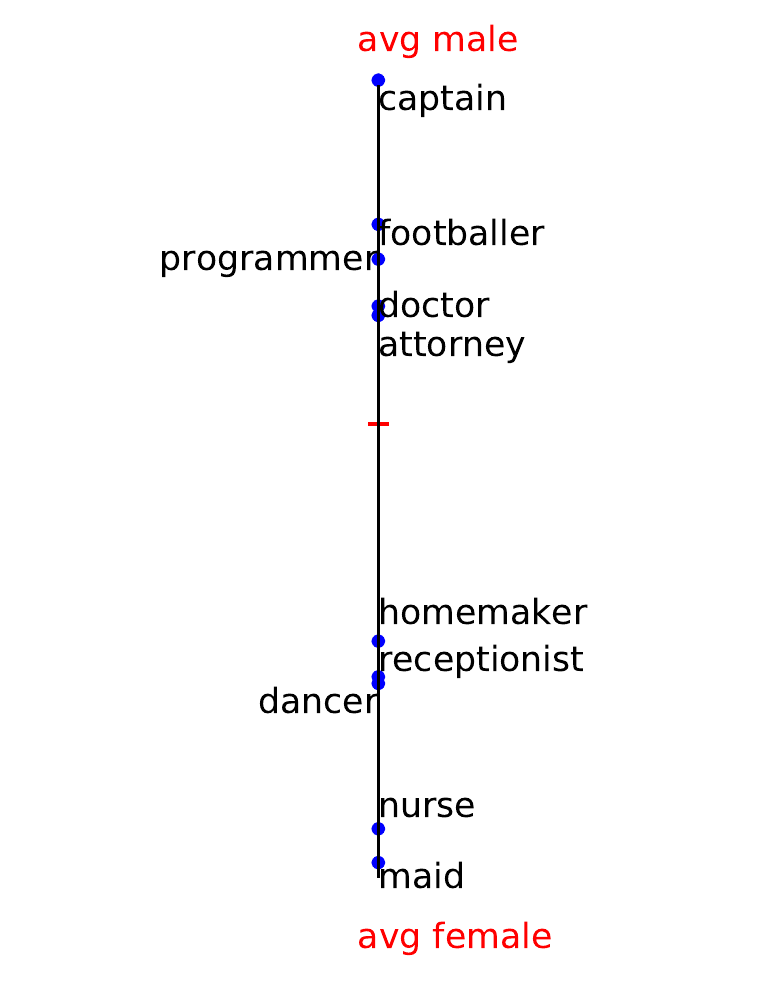}\\
(b) \includegraphics[width=0.2\textwidth,height=0.25\textwidth]{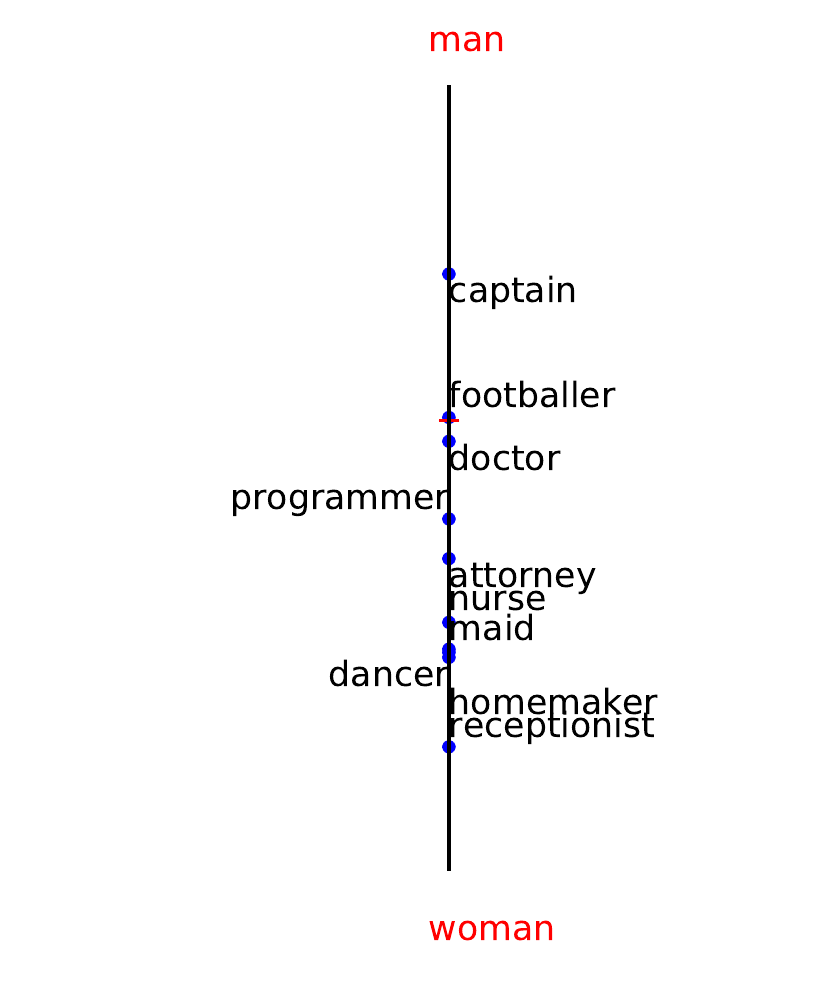}
\includegraphics[width=0.2\textwidth, height=0.25\textwidth]{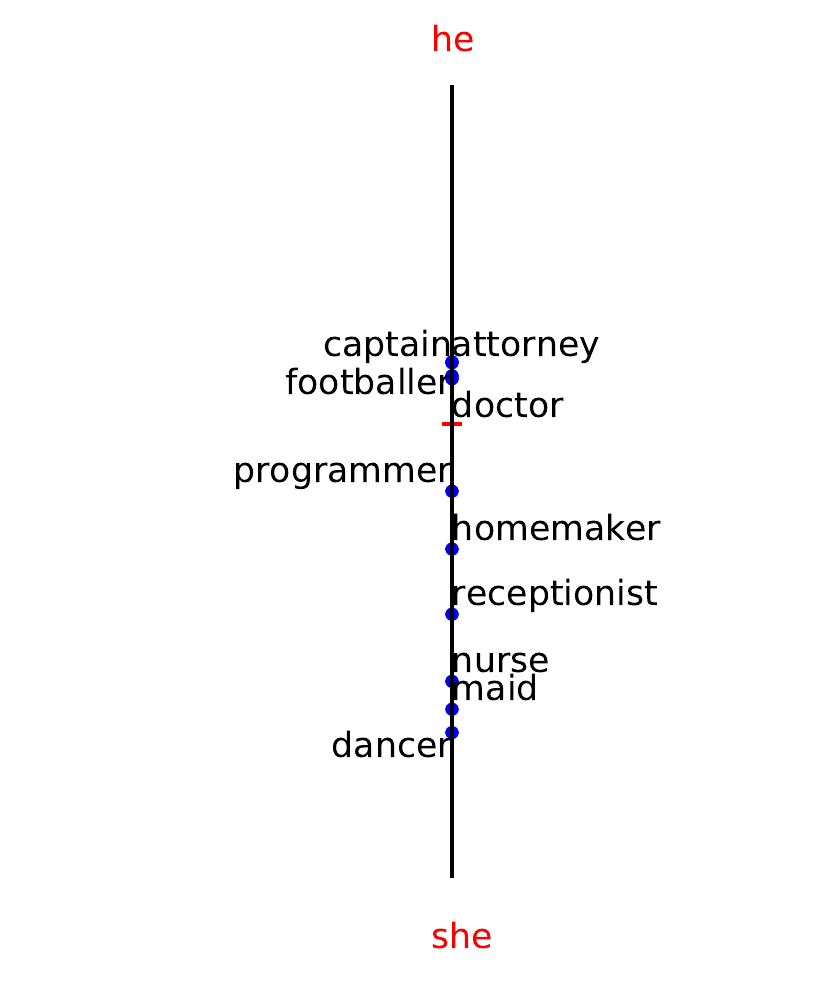}
\includegraphics[width=0.2\textwidth, height=0.25\textwidth]{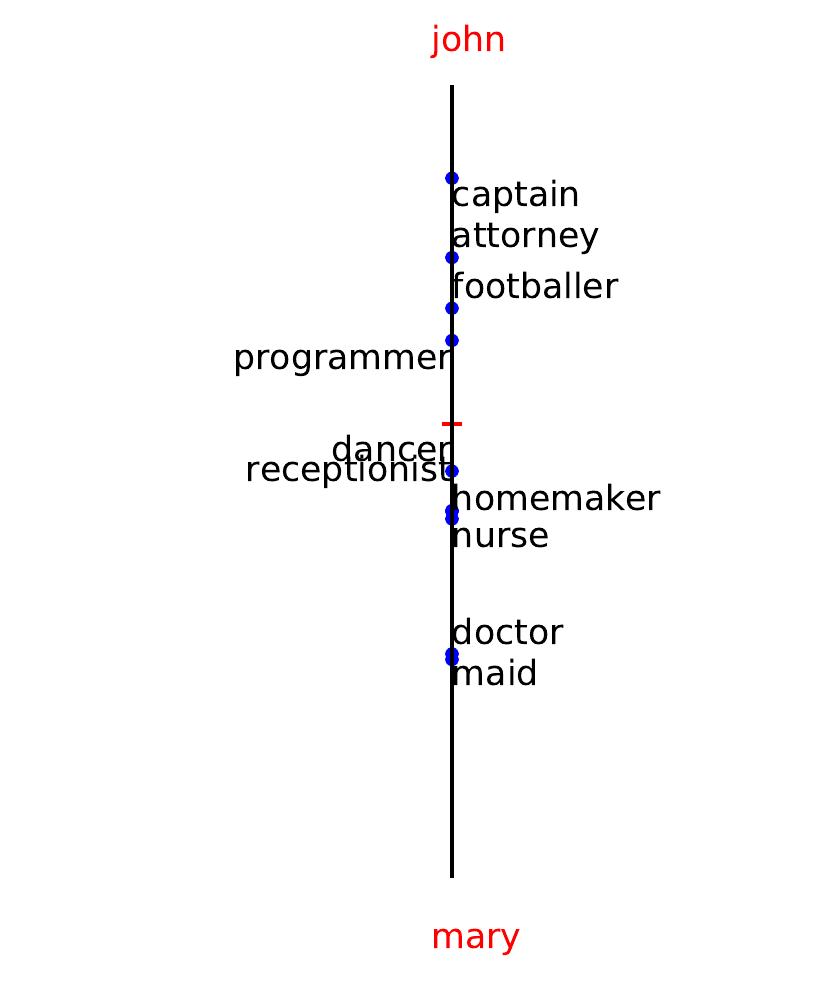}
\includegraphics[width=0.2\textwidth, height=0.25\textwidth]{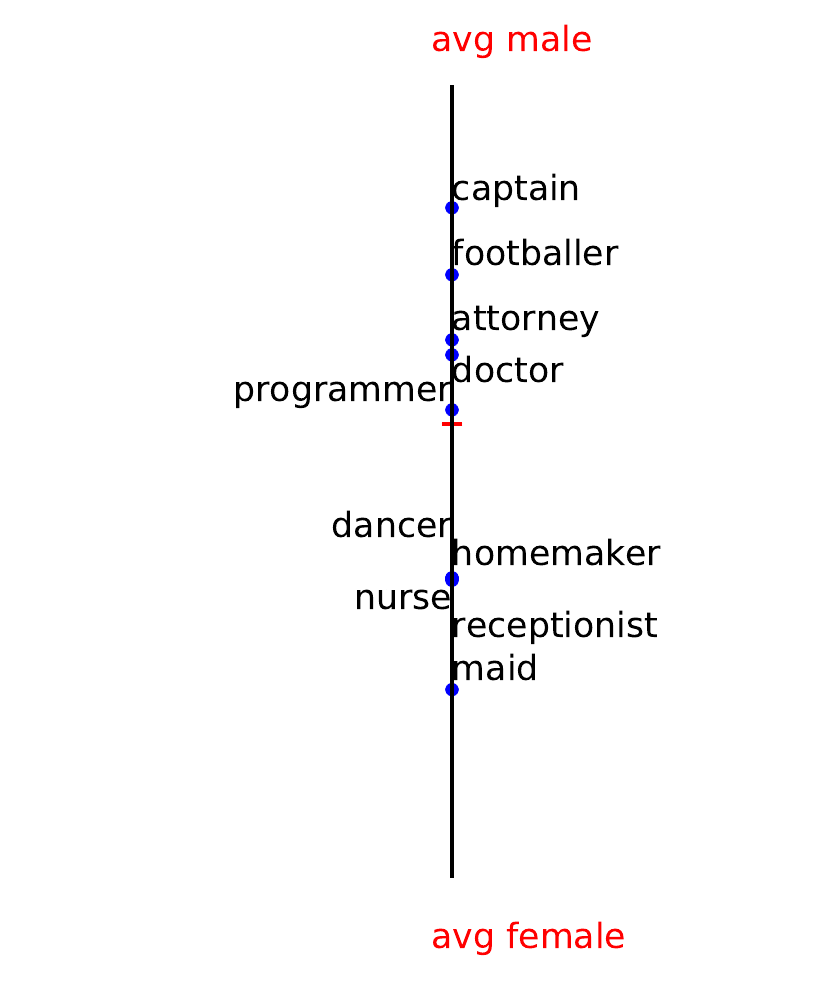}\\
    (c) \includegraphics[width=0.2\textwidth,height=0.25\textwidth]{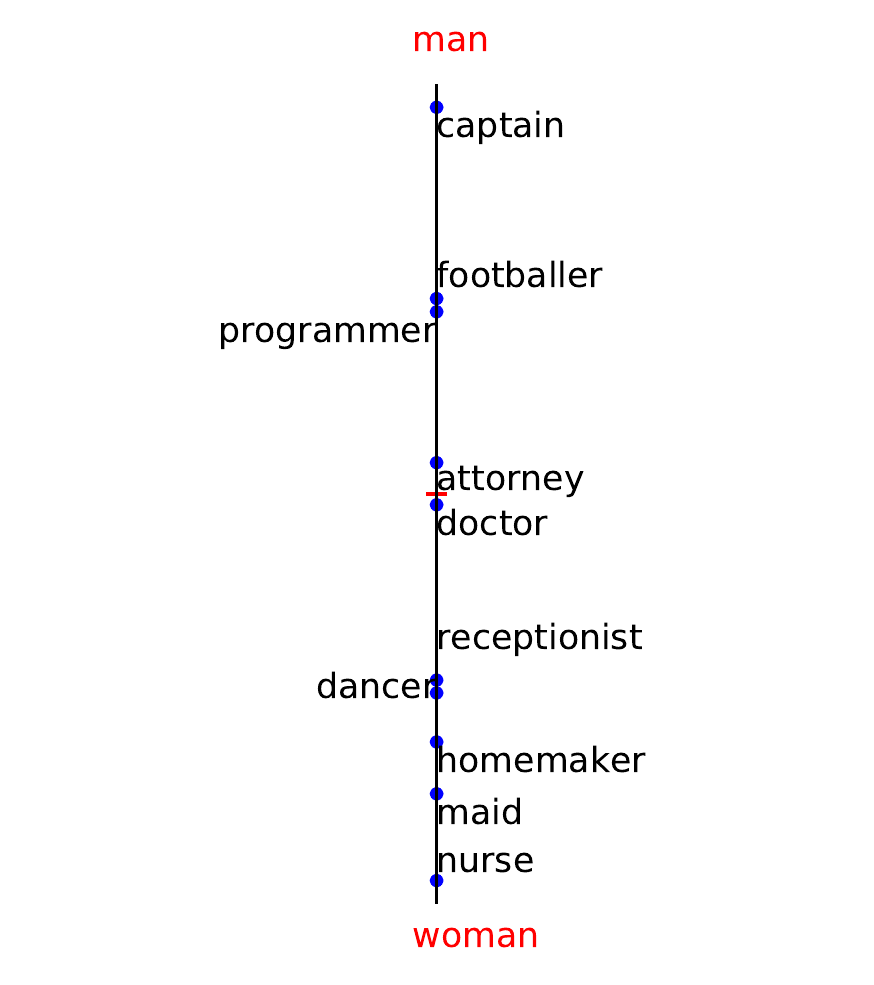}
\includegraphics[width=0.2\textwidth, height=0.25\textwidth]{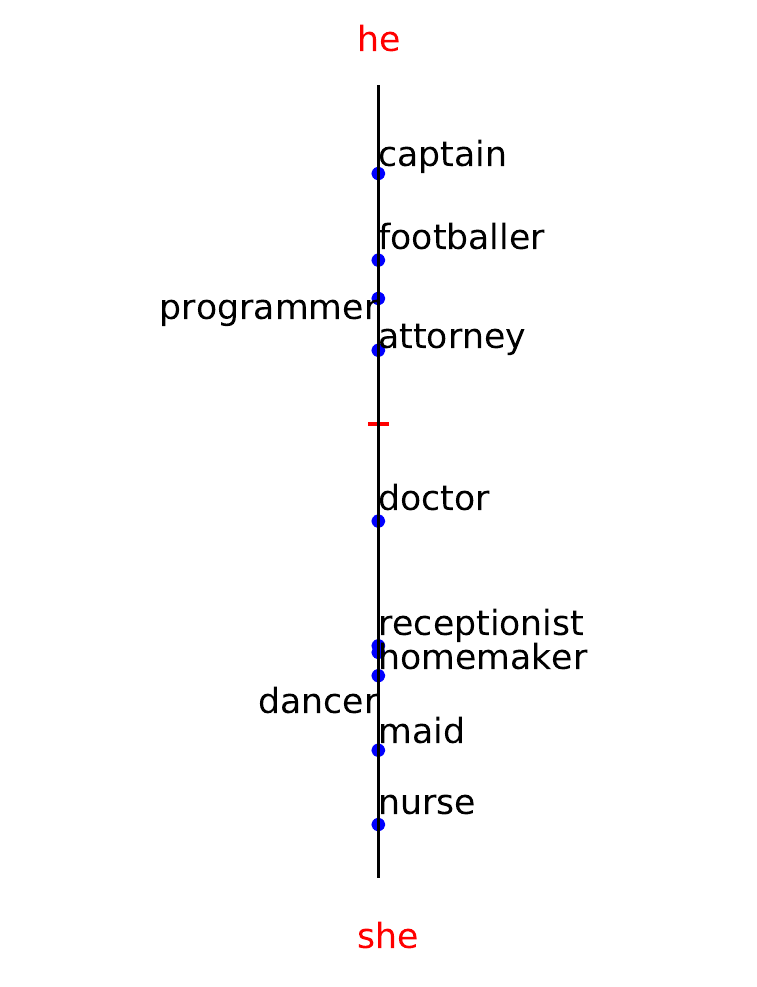}
\includegraphics[width=0.2\textwidth, height=0.25\textwidth]{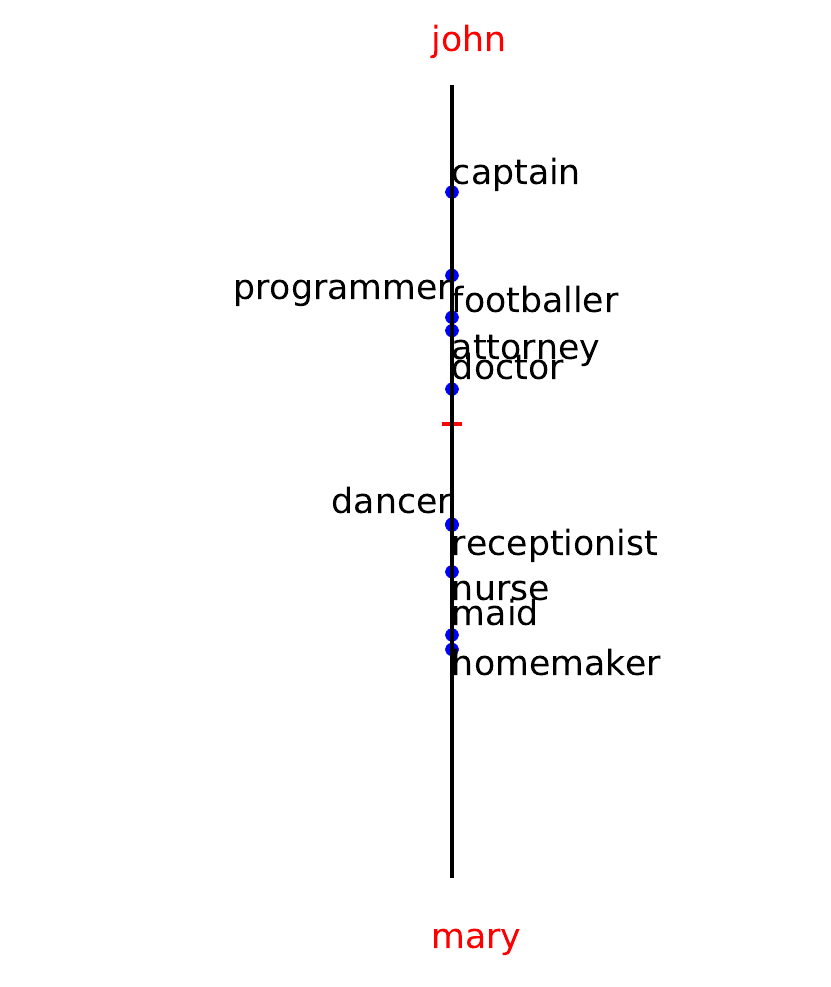}
\includegraphics[width=0.2\textwidth, height=0.25\textwidth]{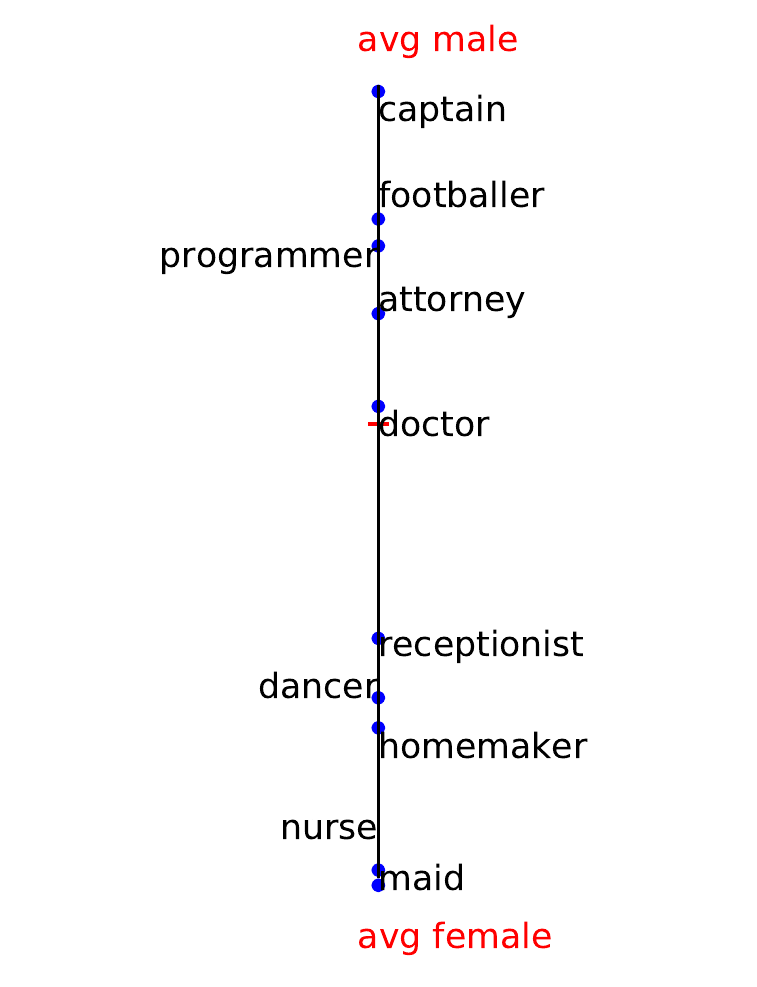}\\

   (d) \includegraphics[width=0.2\textwidth,height=0.25\textwidth]{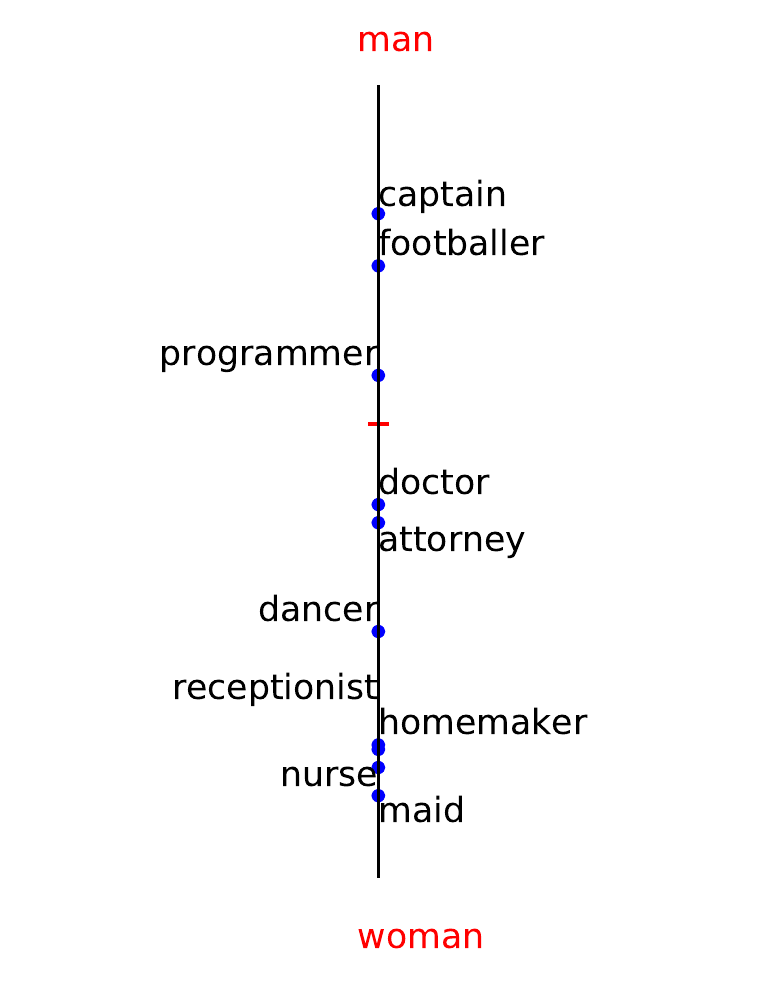}
\includegraphics[width=0.2\textwidth, height=0.25\textwidth]{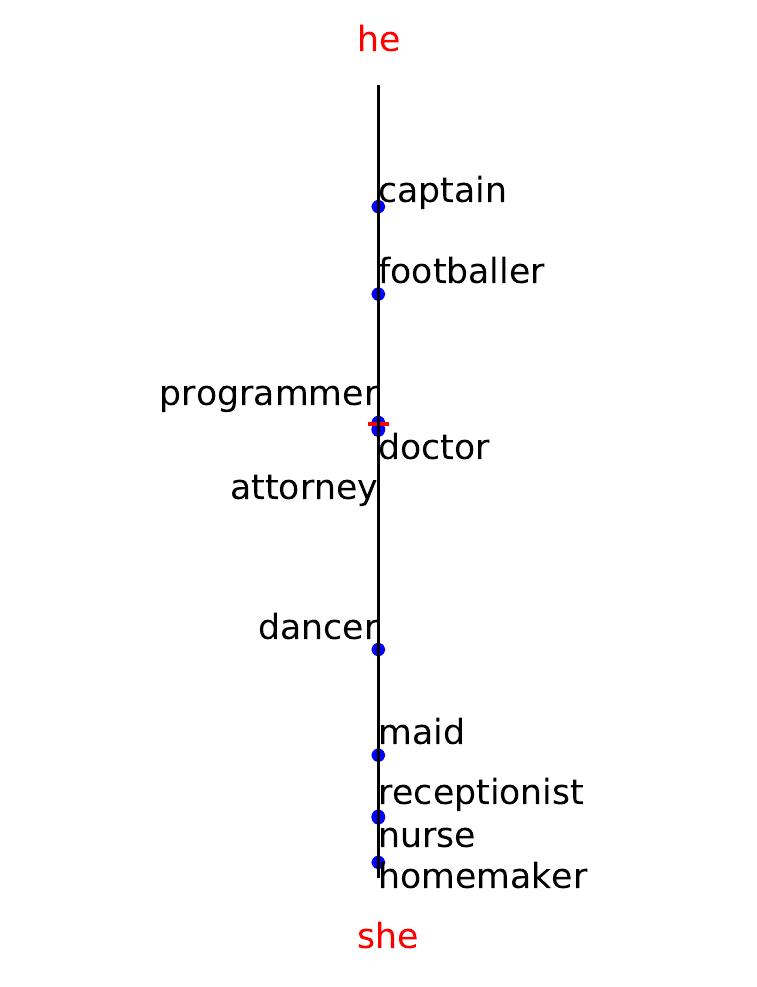}
\includegraphics[width=0.2\textwidth, height=0.25\textwidth]{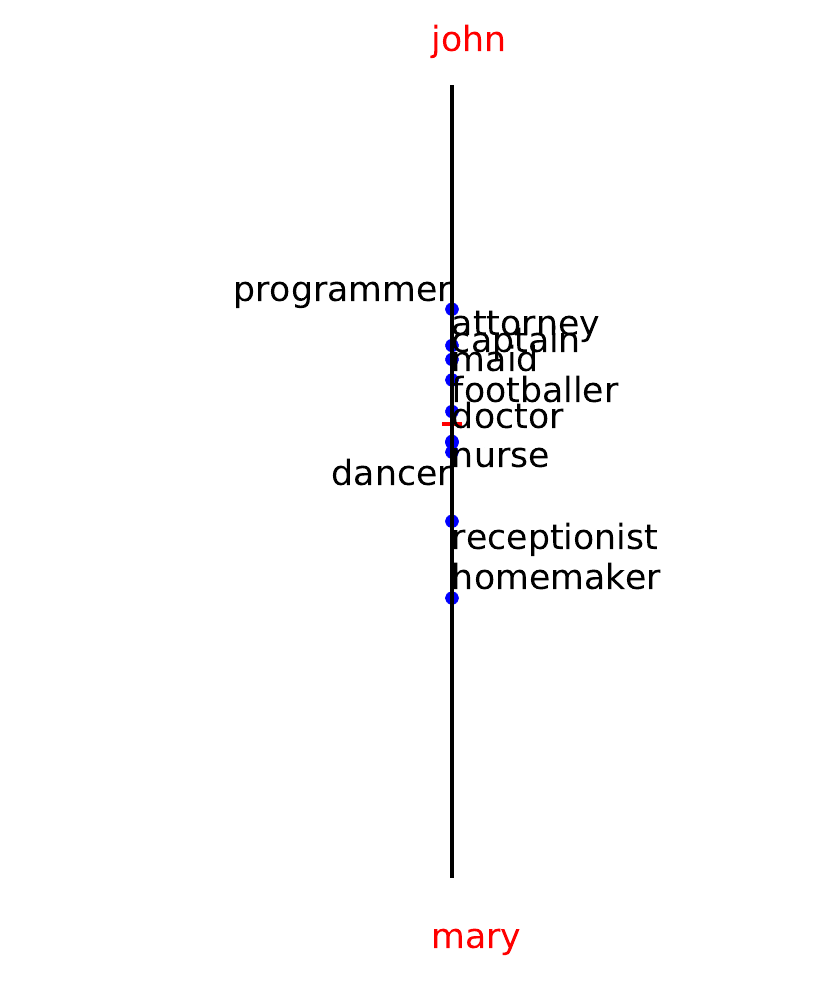}
\includegraphics[width=0.2\textwidth, height=0.25\textwidth]{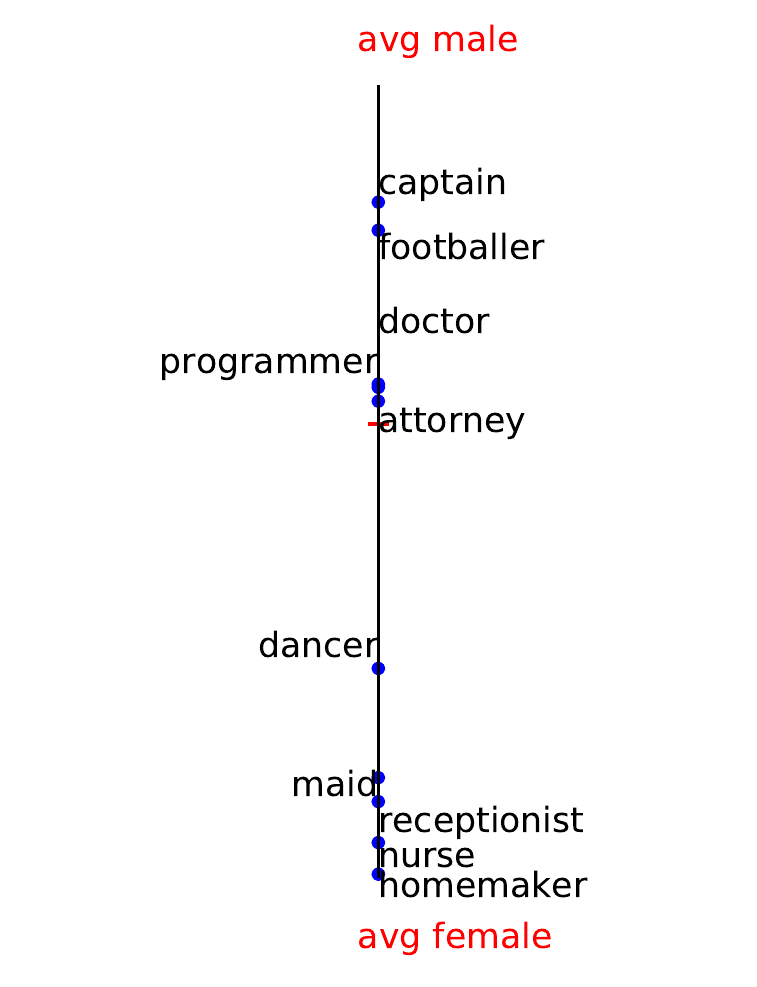}

\caption{Gender Bias in Different Embeddings : (a) \GloVe on Wikipedia, (b) \GloVe on Twitter, (c) \GloVe on Common Crawl and (d) \WordToVec on GoogleNews datasets.}
\label{fig : bias in embeddings}
\end{figure*}

We explore here how gender bias is expressed across different embeddings, datasets and embedding mechanisms. Similar patterns are reflected across all as seen in Figure \ref{fig : bias in embeddings}.

For this verification of the permeative nature of bias across datasets and embeddings, we use the $\GloVe$ embeddings of a Wikipedia dump (\url{dumps.wikimedia.org/enwiki/latest/enwiki-latest-pages-articles.xml.bz2}, $4.7$B tokens) Common Crawl ($840$B tokens, $2.2$M vocab) and Twitter ($27$B tokens, $1.2$M vocab) from \url{https://nlp.stanford.edu/projects/glove/}, and the $\WordToVec$ embedding of Google News ($100$B tokens, $3$M vocab) from  \url{https://code.google.com/archive/p/word2vec/}. 

\begin{figure*}
	\centering
	(a) \includegraphics[width=0.2\textwidth,height=0.25\textwidth]{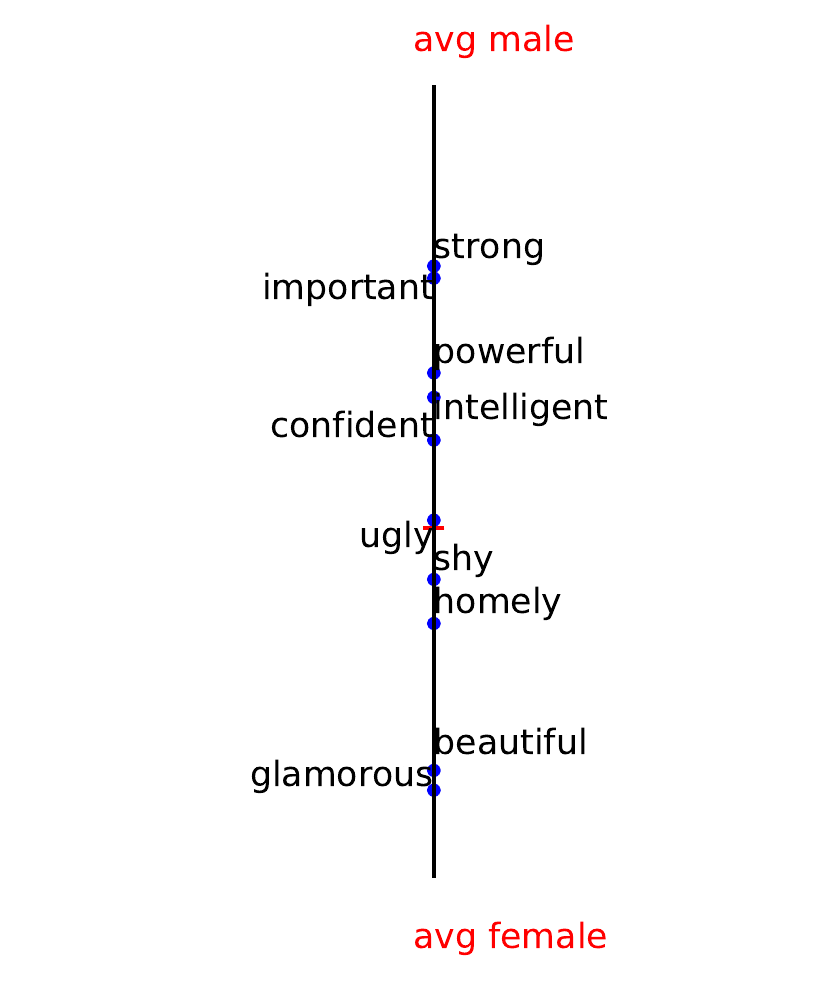}
	(b) \includegraphics[width=0.2\textwidth, height=0.25\textwidth]{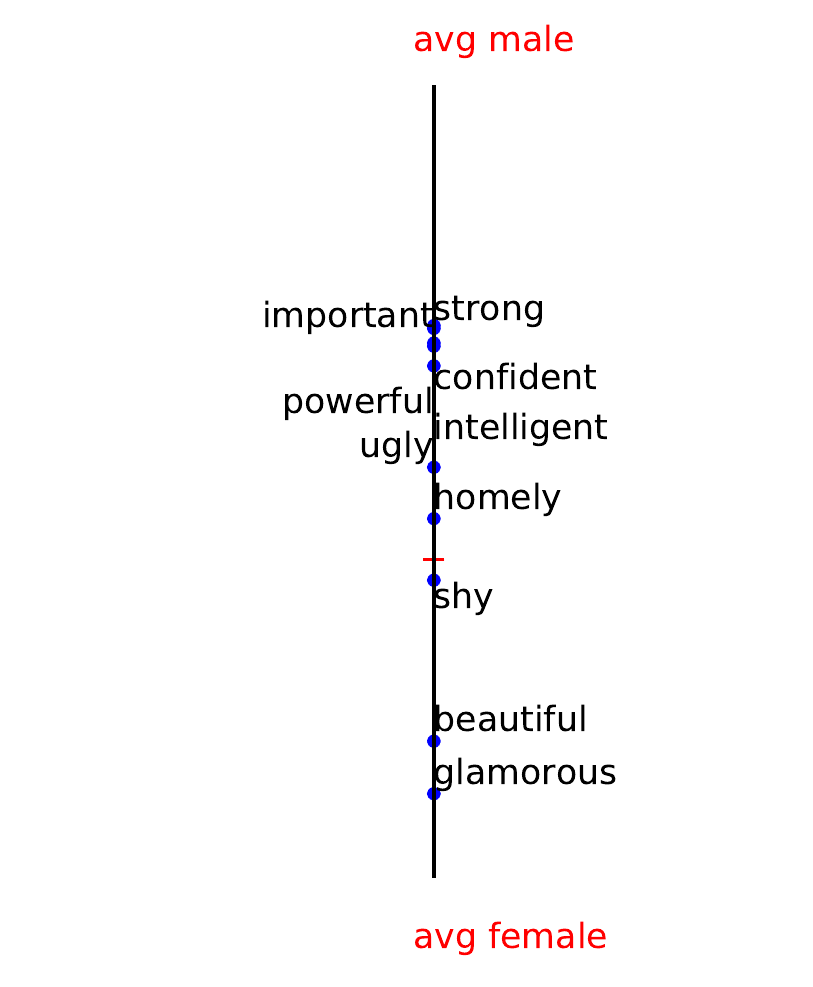}
	(c) \includegraphics[width=0.2\textwidth, height=0.25\textwidth]{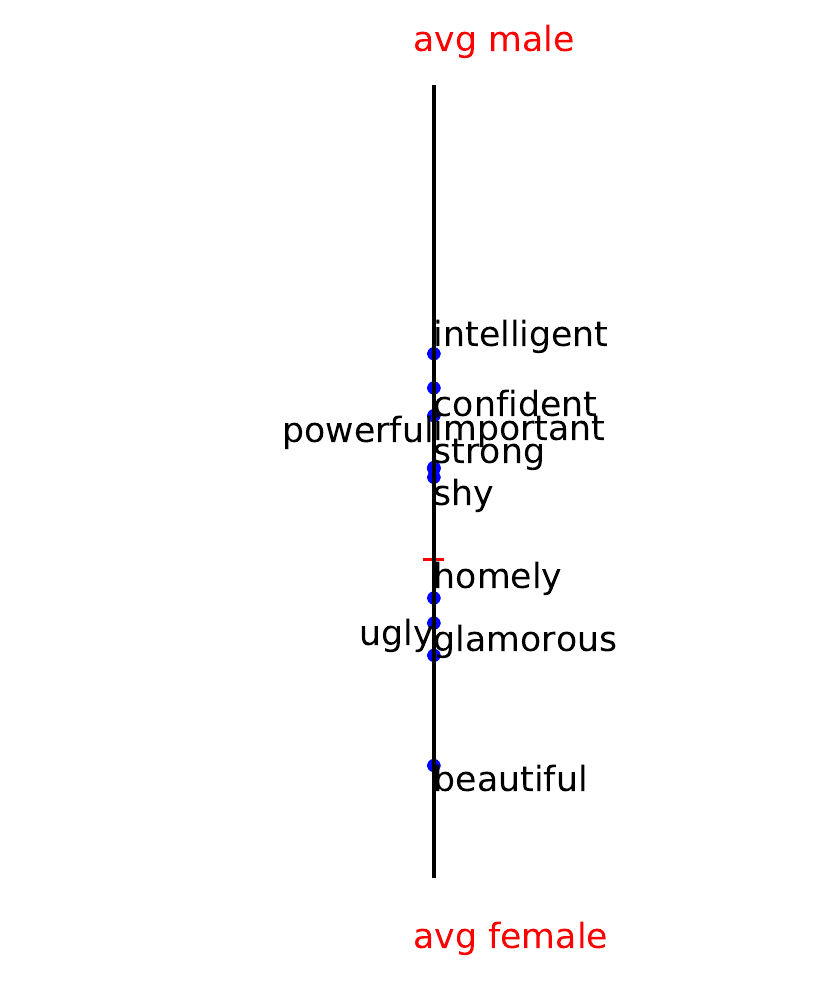}
	(d) \includegraphics[width=0.2\textwidth, height=0.25\textwidth]{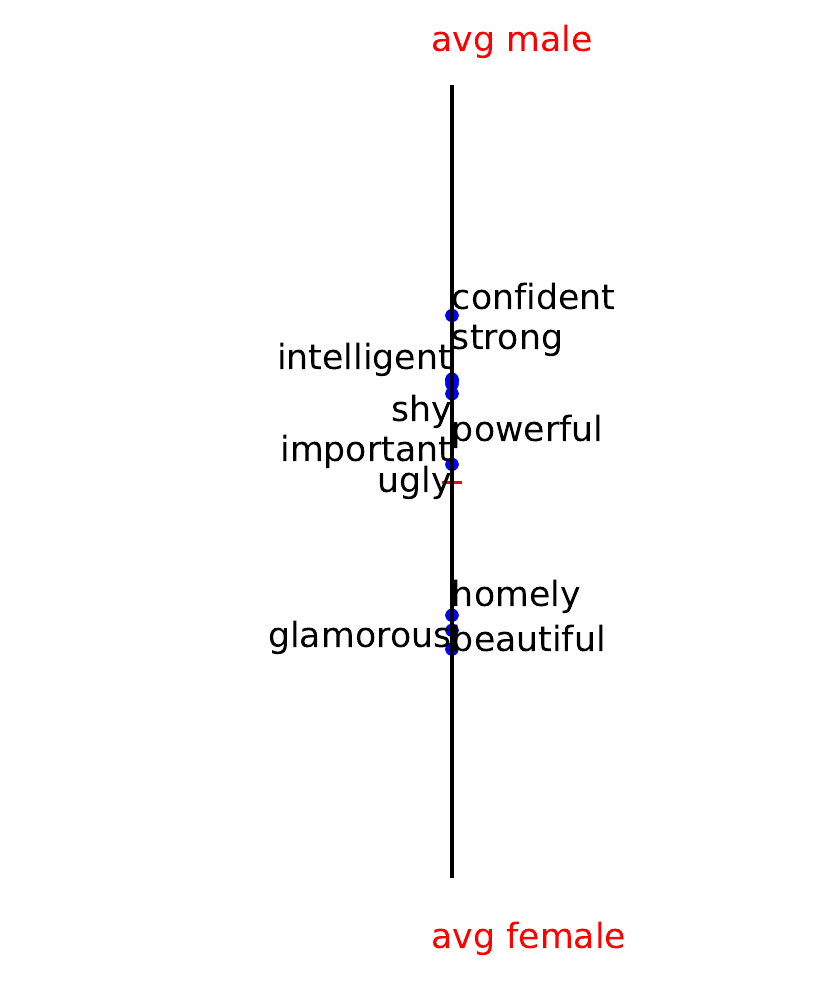}
	
	\caption{Bias in adjectives along the gender direction : (a) $\GloVe$ on default Wikipedia dataset, (b) $\GloVe$ on Common Crawl (840B token dataset), (c) $\GloVe$ on Twitter dataset and (d) $\WordToVec$ on Google News dataset}
	\label{fig : bias adj}
\end{figure*}

\begin{figure}
	\centering
	(a) \includegraphics[width=0.2\textwidth, height = 0.25 \textheight]{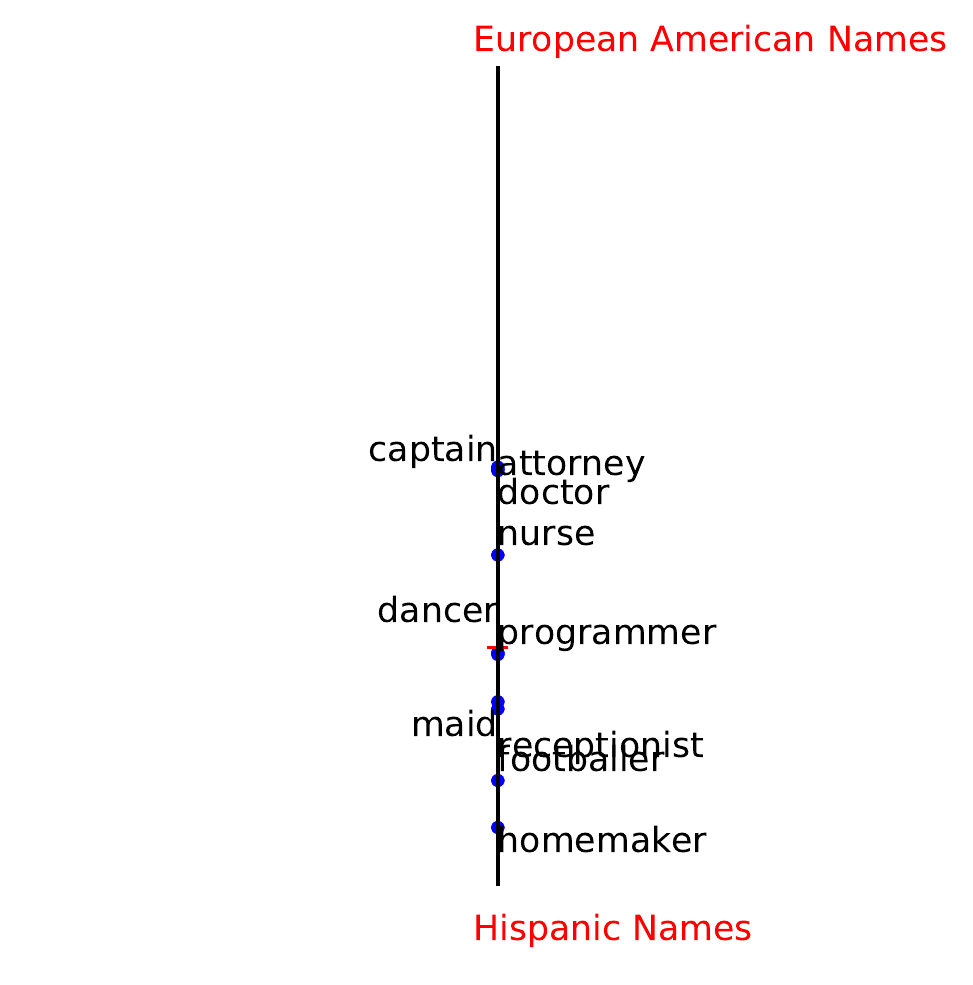}
	(b) \includegraphics[width=0.2\textwidth, height = 0.25 \textheight]{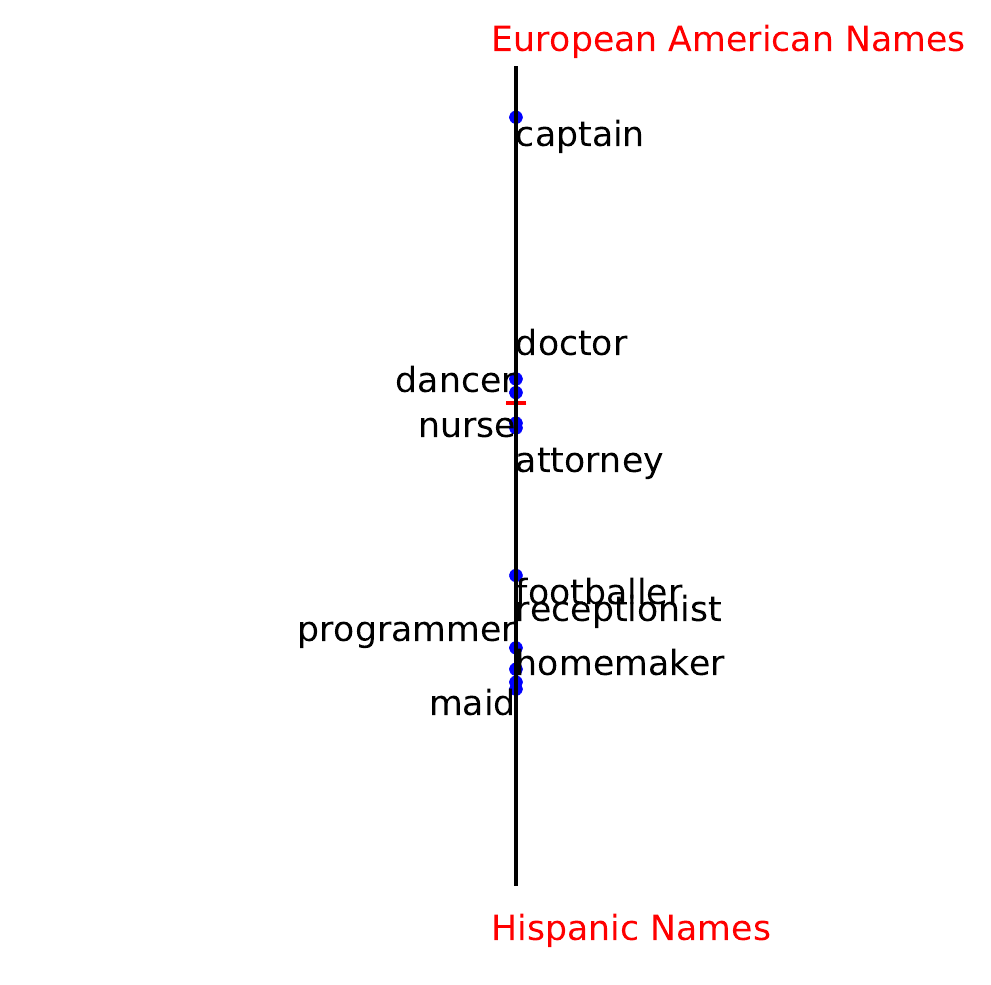} \\
	(c) \includegraphics[width=0.2\textwidth, height = 0.25 \textheight]{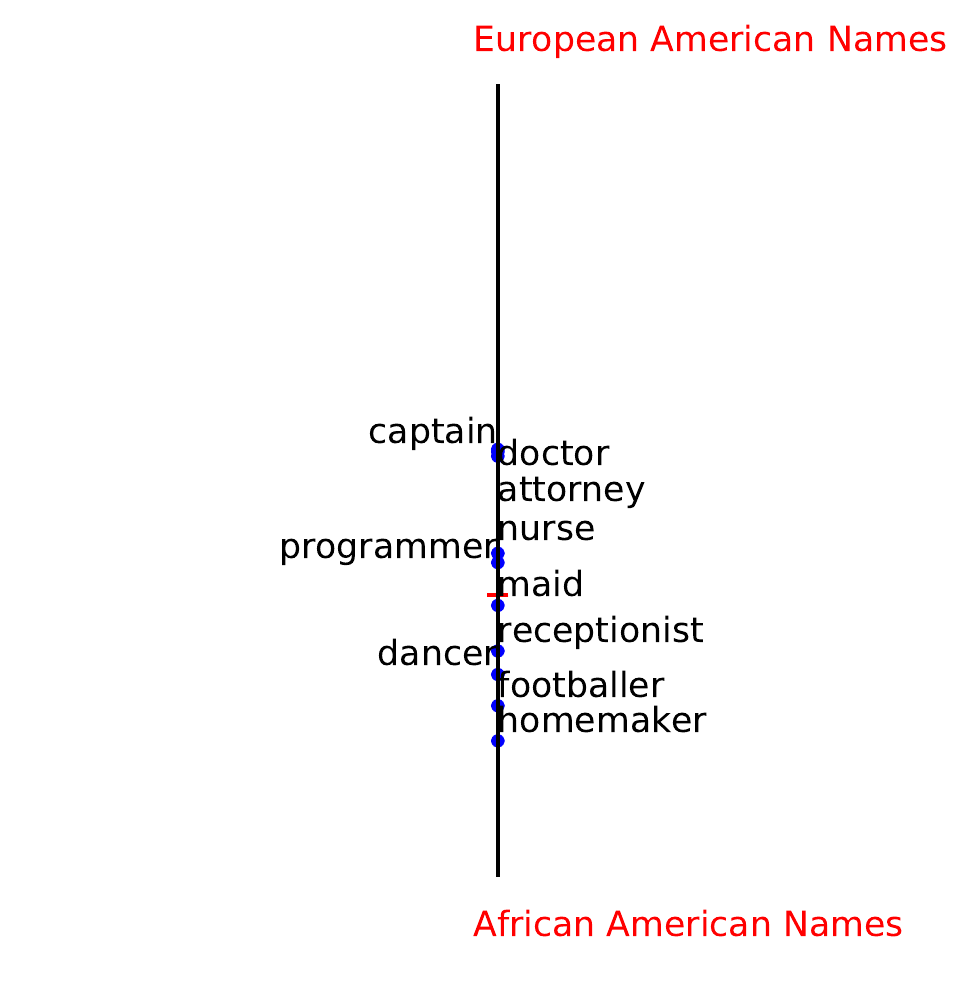}
	(d) \includegraphics[width=0.2\textwidth, height = 0.25 \textheight]{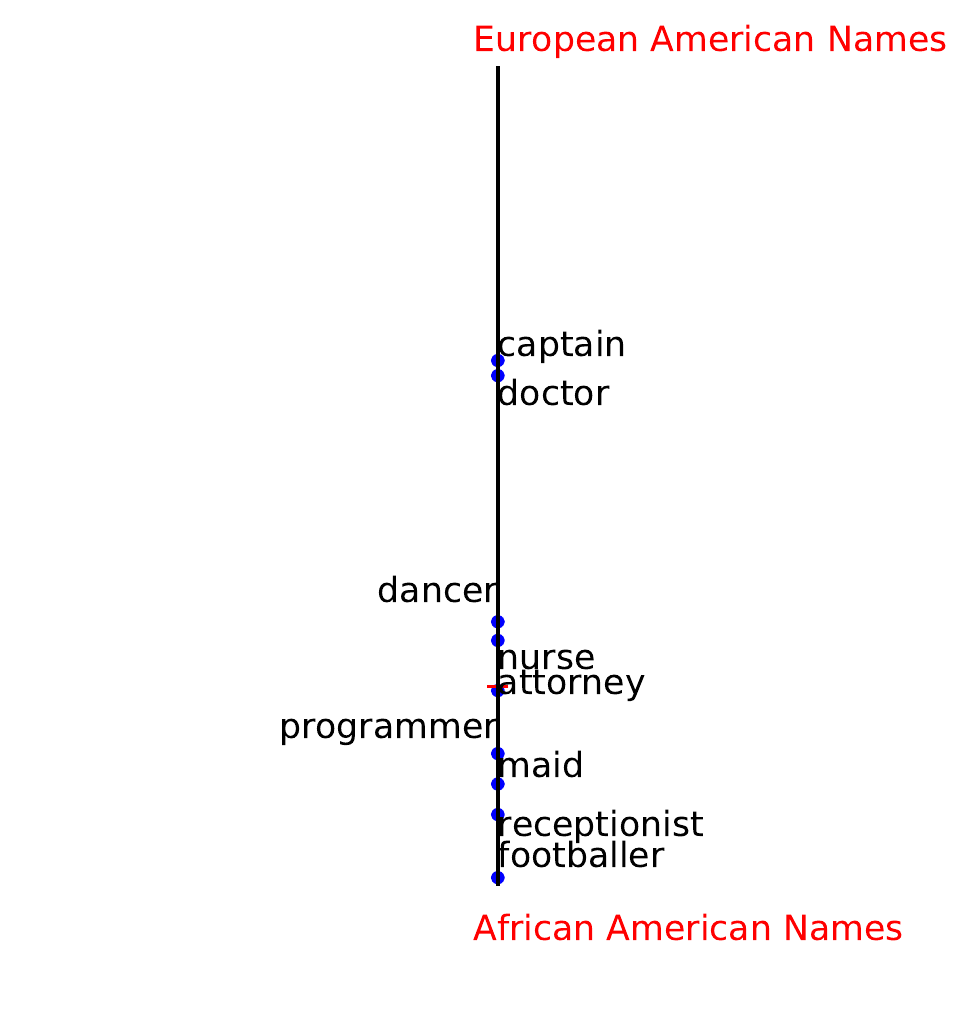}
	\caption{Racial bias in different embeddings : Occupation words along the European American - Hispanic axis in $\GloVe$ embeddings of (a) Common Crawl and (b) Twitter Dataset and the European American - African American axis in $\GloVe$ embeddings of (a) Common Crawl and (b) Twitter Dataset}
	\label{fig:my_label}
\end{figure}

\section{WORD EMBEDDING ASSOCIATION TEST}
\label{app : weat}

Word Embedding Association Test (WEAT) was defined as an analogue to Implicit Association Test (IAT) by Caliskan \etal \cite{Caliskan183}. It checks for human like bias associated with words in word embeddings. For example, it found career oriented words (executive, career, etc) more associated with male names and male gendered words ('man','boy' etc) than female names and gendered words and family oriented words ('family','home' etc) more associated with female names and words than male. We list a set of words used for WEAT by Calisan \etal and that we used in our work below.

For two sets of target words X and Y and attribute words A and B, the WEAT test statistic is :

$s(X,Y,A,B) = \sum_{x\in X} s(x,A,B) - s(y,A,B)$

where, 

$s(w,A,B) = mean_{a \in A} cos(a,w) - mean_{b \in B} cos(b,w)$ and, $cos(a,b)$ is the cosine distance between vector a and b.

This score is normalized by $std-dev_{w \in X \cup Y} s(w,A,B)$. So, closer to 0 this value is, the less bias or preferential association target word groups have to the attribute word groups.

Here target words are occupation words or career/family oriented words and attributes are male/female words or names.

Career : \{ executive, management, professional, corporation, salary, office, business, career \} 

Family : \{ home, parents, children, family, cousins, marriage, wedding, relatives \}

Male names : \{ john, paul, mike, kevin, steve, greg, jeff, bill \} 

Female names : \{ amy, joan, lisa, sarah, diana, kate, ann, donna \} 

Male words : \{ male, man, boy, brother, he, him, his, son \} 

Female words : \{ female, woman, girl, she, her, hers, daughter \} \\

\section{DETECTING THE GENDER DIRECTION}
\label{app : gender dir}
\begin{figure}
	\centering
	\includegraphics[width=0.15\textwidth]{gender1.pdf}
	\includegraphics[width=0.15\textwidth]{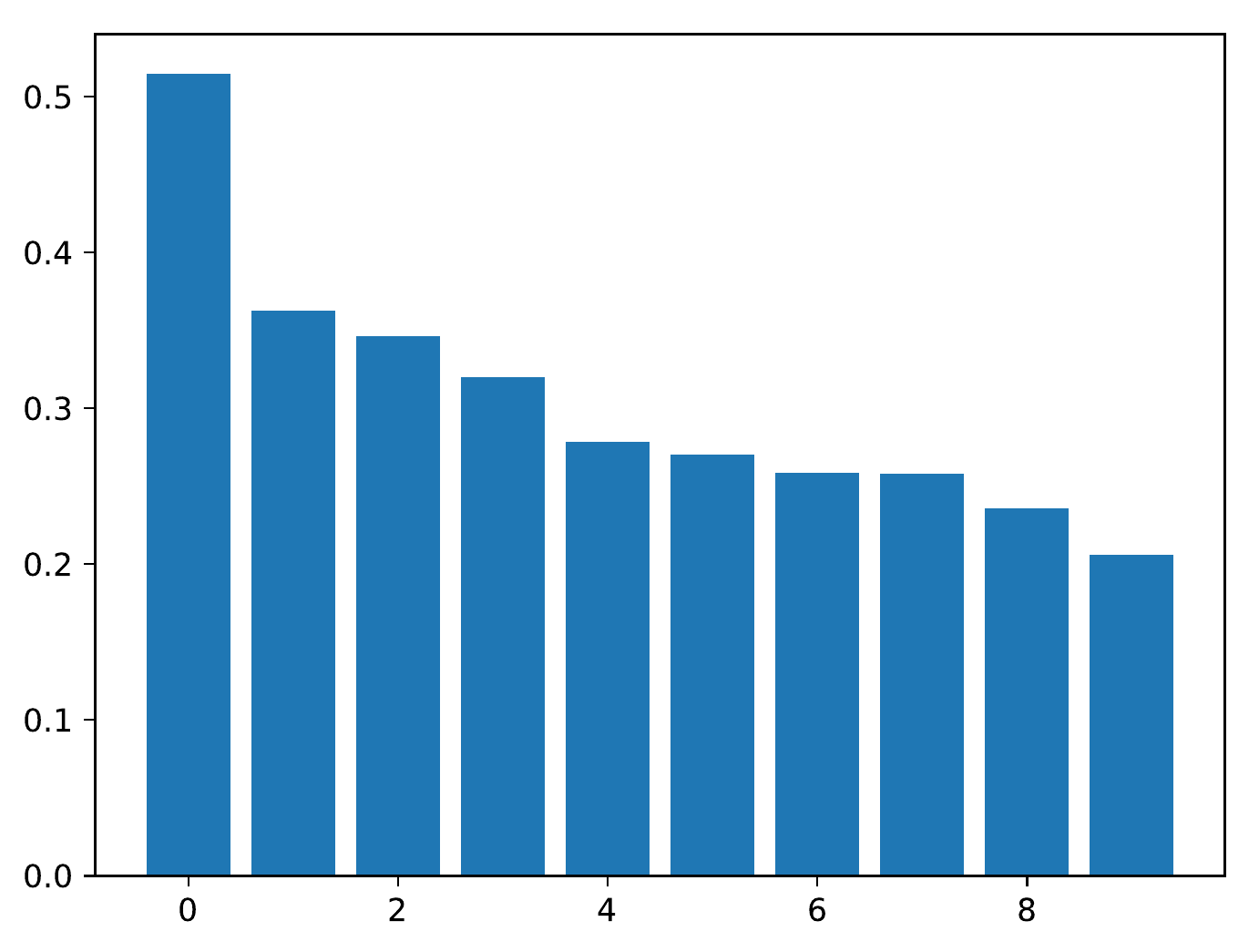} \\
	\includegraphics[width=0.15\textwidth]{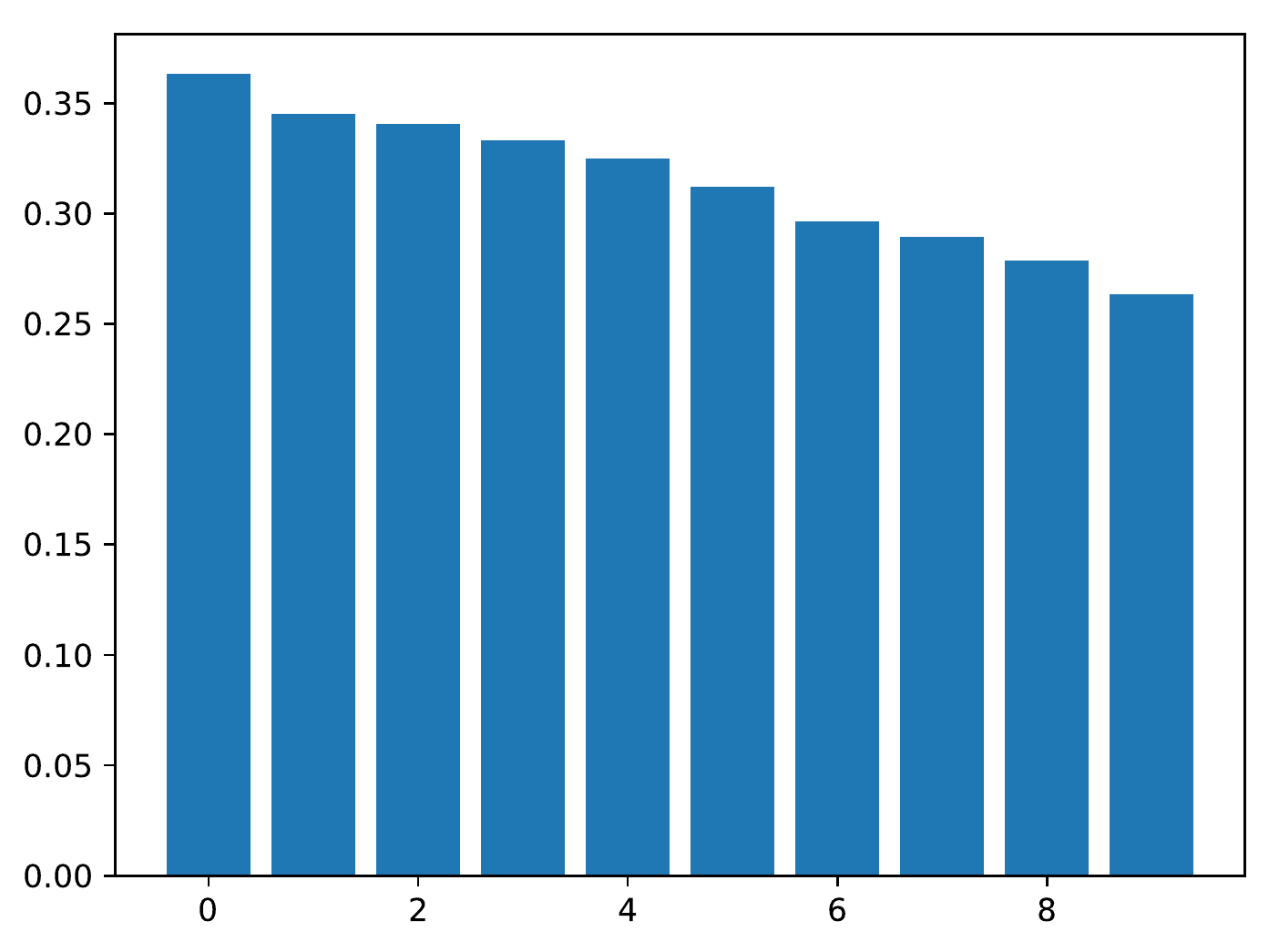}
	\includegraphics[width=0.15\textwidth]{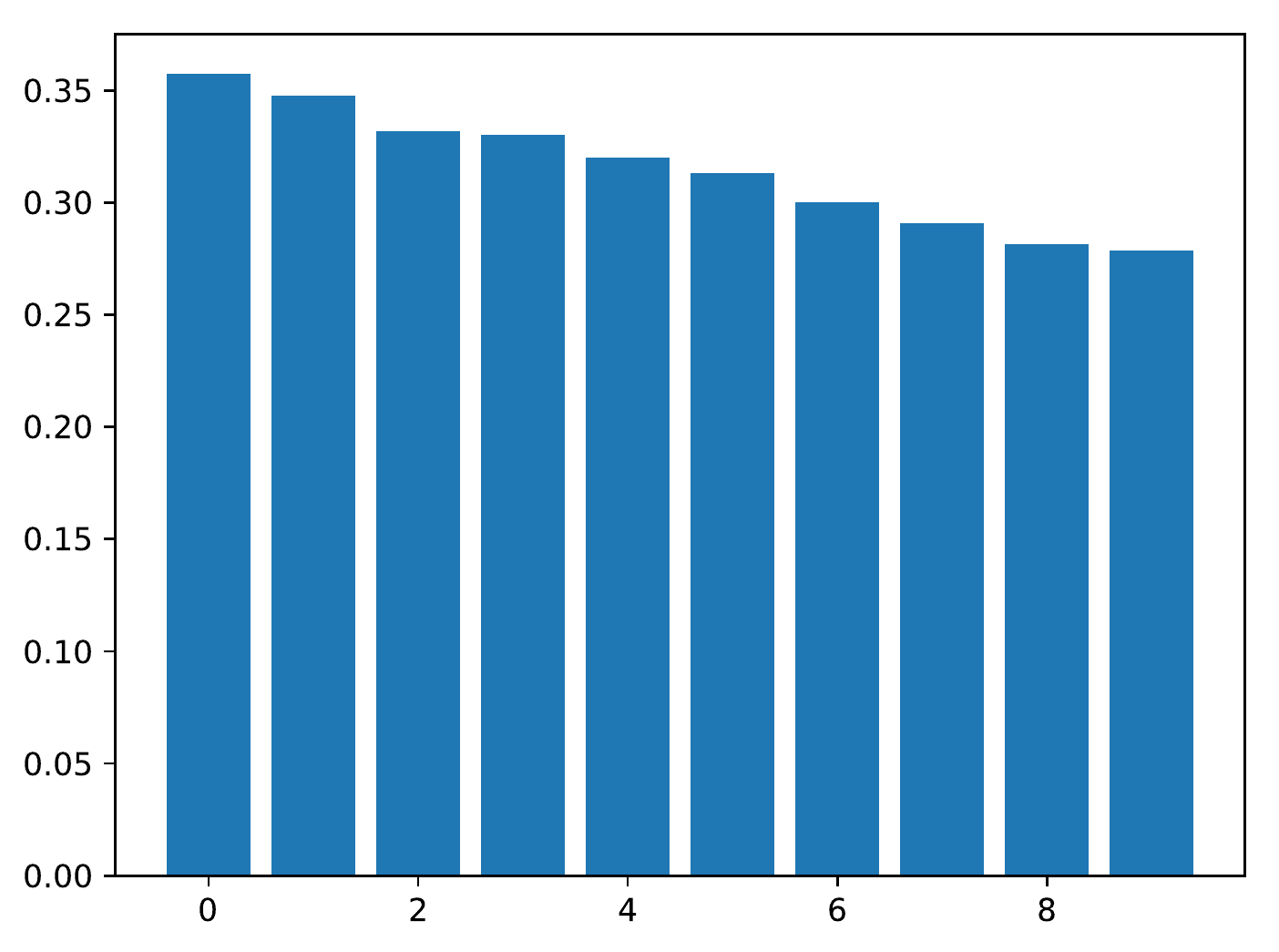}
	
	\caption{Fractional singular values for (a) male-female word pairs (b) one gendered word - one random word (c) random word pair (d) random unit vectors}
	\label{fig:SVDcharts}
\end{figure}

For this, we take a set of gendered word pairs as listed in Table \ref{tbl : Gendered Words}.
From our default Wikipedia dataset, using the embedded vectors for these word pairs (i.e., \wv{woman}{man}, \wv{she}{he}, etc), we create a basis for the subspace $F$, of dimension $10$.   We then try to understand the distribution of variance in this subspace.  To do so, we project the entire dataset onto this subspace $F$, and take the SVD.  The top chart in Figure \ref{fig:SVDcharts} shows the singular values of the entire data in this subspace $F$.  We observe that there is a dominant first singular vector/value which is almost twice the size of the second value.  After the this drop, the decay is significantly more gradual.  This suggests to use only the top singular vector of $F$ as the gender subspace, not $2$ or more of these vectors.   

To grasp how much of this variation is from the correlation along the gender direction, and how much is just random variation, we repeat this experiment, again in Figure \ref{fig:SVDcharts}, with different ways of creating the subspace $F$.  
First in chart (b), we generate $10$ vectors, with one word chosen randomly, and one chosen from the gendered set (e.g., chair-woman).  
Second in chart (c), we generate $10$ vectors between two random words from our set of the $100{,}000$ most frequent words; these are averaged over $100$ random iterations due to higher variance in the plots.  
Finally in chart (d), we generate $10$ random unit vectors in $\b{R}^{300}$.    
We observe that the pairs with one gendered vector in each pair still exhibits a significant drop in singular values, but not as drastic as with both pairs. The other two approaches have no significant drop since they do not in general contain a gendered word with interesting subspace. All remaining singular values, and their decay appears similar to the non-leading ones from the charts (a) and (b). This further indicates that there is roughly one important gender direction, and any related subspace is not significantly different than a random one in the word-vector embedding.

%
%

Now, for any word $w$ in vocabulary $W$ of the embedding, we can define $w_B$ as the part of $w$ along the gender direction. 


Based on the experiments shown in Figure \ref{fig:SVDcharts}, it is justified to take the gender direction as the (normalized) first right singular vector, $v_B$, or the full data set data projected onto the subspace $F$.  Then, the component of a word vector $w$ along $v_B$ is simply $\langle w, v_B\rangle v_B$. 

Calculating this component when the gender subspace is defined by two or more of the top right singular vectors of $V$ can be done similarly.

We should note here that the gender subspace defined here passes through the origin. Centering the data and using PCA to define the gender subspace lets the gender subspace not pass through the origin. We see a comparison in the two methods in Section \ref{sec : tests} as $HD$ uses PCA and we use SVD to define the gender direction.

\vspace{2mm}

\section{Word Lists}
\label{app : word lists}
\subsection{Word Pairs used for Flipping}
\label{app : flipping}
actor actress\\
author authoress\\
bachelor spinster\\
boy girl\\
brave squaw\\
bridegroom bride\\
brother sister\\
conductor conductress\\
count countess\\
czar czarina\\
dad mum\\
daddy mummy\\
duke duchess\\
emperor empress\\
father mother\\
father-in-law mother-in-law\\
fiance fiancee\\
gentleman lady\\
giant giantess\\
god goddess\\
governor matron\\
grandfather grandmother\\
grandson granddaughter\\
he she\\
headmaster headmistress\\
heir heiress\\
hero heroine\\
him her\\
himself herself\\
host hostess\\
hunter huntress\\
husband wife\\
king queen\\
lad lass\\
landlord landlady\\
lord lady\\
male female\\
man woman\\
manager manageress\\
manservant maidservant\\
masseur masseuse\\
master mistress\\
mayor mayoress\\
milkman milkmaid\\
millionaire millionairess\\
monitor monitress\\
monk nun\\
mr mrs\\
murderer murderess\\
nephew niece\\
papa mama\\
poet poetess\\
policeman policewoman\\
postman postwoman\\
postmaster postmistress\\
priest priestess\\
prince princess\\
prophet prophetess\\
proprietor proprietress\\\
shepherd shepherdess\\
sir madam\\
son daughter\\
son-in-law daughter-in-law\\
step-father step-mother\\
step-son step-daughter\\
steward stewardess\\
sultan sultana\\
tailor tailoress\\
uncle aunt\\
usher usherette\\
waiter waitress\\
washerman washerwoman\\
widower widow\\
wizard witch\\

\subsection{Occupation Words}
\label{sec : occupation words}
detective\\
ambassador\\
coach\\
officer\\
epidemiologist\\
rabbi\\
ballplayer\\
secretary\\
actress\\
manager\\
scientist\\
cardiologist\\
actor\\
industrialist\\
welder\\
biologist\\
undersecretary\\
captain\\
economist\\
politician\\
baron\\
pollster\\
environmentalist\\
photographer\\
mediator\\
character\\
housewife\\
jeweler\\
physicist\\
hitman\\
geologist\\
painter\\
employee\\
stockbroker\\
footballer\\
tycoon\\
dad\\
patrolman\\
chancellor\\
advocate\\
bureaucrat\\
strategist\\
pathologist\\
psychologist\\
campaigner\\
magistrate\\
judge\\
illustrator\\
surgeon\\
nurse\\
missionary\\
stylist\\
solicitor\\
scholar\\
naturalist\\
artist\\
mathematician\\
businesswoman\\
investigator\\
curator\\
soloist\\
servant\\
broadcaster\\
fisherman\\
landlord\\
housekeeper\\
crooner\\
archaeologist\\
teenager\\
councilman\\
attorney\\
choreographer\\
principal\\
parishioner\\
therapist\\
administrator\\
skipper\\
aide\\
chef\\
gangster\\
astronomer\\
educator\\
lawyer\\
midfielder\\
evangelist\\
novelist\\
senator\\
collector\\
goalkeeper\\
singer\\
acquaintance\\
preacher\\
trumpeter\\
colonel\\
trooper\\
understudy\\
paralegal\\
philosopher\\
councilor\\
violinist\\
priest\\
cellist\\
hooker\\
jurist\\
commentator\\
gardener\\
journalist\\
warrior\\
cameraman\\
wrestler\\
hairdresser\\
lawmaker\\
psychiatrist\\
clerk\\
writer\\
handyman\\
broker\\
boss\\
lieutenant\\
neurosurgeon\\
protagonist\\
sculptor\\
nanny\\
teacher\\
homemaker\\
cop\\
planner\\
laborer\\
programmer\\
philanthropist\\
waiter\\
barrister\\
trader\\
swimmer\\
adventurer\\
monk\\
bookkeeper\\
radiologist\\
columnist\\
banker\\
neurologist\\
barber\\
policeman\\
assassin\\
marshal\\
waitress\\
artiste\\
playwright\\
electrician\\
student\\
deputy\\
researcher\\
caretaker\\
ranger\\
lyricist\\
entrepreneur\\
sailor\\
dancer\\
composer\\
president\\
dean\\
comic\\
medic\\
legislator\\
salesman\\
observer\\
pundit\\
maid\\
archbishop\\
firefighter\\
vocalist\\
tutor\\
proprietor\\
restaurateur\\
editor\\
saint\\
butler\\
prosecutor\\
sergeant\\
realtor\\
commissioner\\
narrator\\
conductor\\
historian\\
citizen\\
worker\\
pastor\\
serviceman\\
filmmaker\\
sportswriter\\
poet\\
dentist\\
statesman\\
minister\\
dermatologist\\
technician\\
nun\\
instructor\\
alderman\\
analyst\\
chaplain\\
inventor\\
lifeguard\\
bodyguard\\
bartender\\
surveyor\\
consultant\\
athlete\\
cartoonist\\
negotiator\\
promoter\\
socialite\\
architect\\
mechanic\\
entertainer\\
counselor\\
janitor\\
firebrand\\
sportsman\\
anthropologist\\
performer\\
crusader\\
envoy\\
trucker\\
publicist\\
commander\\
professor\\
critic\\
comedian\\
receptionist\\
financier\\
valedictorian\\
inspector\\
steward\\
confesses\\
bishop\\
shopkeeper\\
ballerina\\
diplomat\\
parliamentarian\\
author\\
sociologist\\
photojournalist\\
guitarist\\
butcher\\
mobster\\
drummer\\
astronaut\\
protester\\
custodian\\
maestro\\
pianist\\
pharmacist\\
chemist\\
pediatrician\\
lecturer\\
foreman\\
cleric\\
musician\\
cabbie\\
fireman\\
farmer\\
headmaster\\
soldier\\
carpenter\\
substitute\\
director\\
cinematographer\\
warden\\
marksman\\
congressman\\
prisoner\\
librarian\\
magician\\
screenwriter\\
provost\\
saxophonist\\
plumber\\
correspondent\\
organist\\
baker\\
doctor\\
constable\\
treasurer\\
superintendent\\
boxer\\
physician\\
infielder\\
businessman\\
protege\\

\subsection{Names used for Gender Bias Detection}
Male : \{ john, william, george, liam, andrew, michael, louis, tony, scott, jackson \} \\
Female : \{ mary, victoria, carolina, maria, anne, kelly, marie, anna, sarah, jane \} \\

\subsection{Names used for Racial Bias Detection and Dampening}
European American : \{ brad, brendan, geoffrey, greg, brett, matthew, neil, todd, nancy, amanda, emily, rachel \} \\
African American : \{ darnell, hakim, jermaine, kareem, jamal, leroy, tyrone, rasheed, yvette, malika, latonya, jasmine \} \\
Hispanic : \{ alejandro, pancho, bernardo, pedro, octavio, rodrigo, ricardo, augusto, carmen, katia, marcella , sofia \} \\

\subsection{Names used for Age related Bias Detection and Dampening}
Aged : \{ ruth, william, horace, mary, susie, amy, john, henry, edward, elizabeth \} \\
Youth : \{ taylor, jamie, daniel, aubrey, alison, miranda, jacob, arthur, aaron, ethan \} \\


\end{document}